\documentclass[preprint,3p]{elsarticle}

\usepackage{amsmath}
\usepackage{amssymb}
\usepackage{stackrel}
\usepackage{amsfonts}
\usepackage{hyperref}
\usepackage{graphicx}
\usepackage{epstopdf}
\usepackage{color}
\usepackage{algorithmic}
\usepackage{algorithm}
\usepackage{subfig}
\usepackage{amssymb}
\usepackage{amsthm}

\newtheorem{lemma}{Lemma}
\newtheorem{theorem}{Theorem}

\journal{Information Sciences}

\begin{document}

\begin{frontmatter}



\title{Average Convergence Rate of Evolutionary Algorithms in Continuous Optimization \tnoteref{ref1} \tnoteref{ref2}}

\author[CHEN]{Yu Chen}
\address[CHEN]{School of Science, Wuhan University of Technology, Wuhan, 430070, China.}
\ead{ychen@whut.edu.cn }

\author[HE]{Jun He\corref{cor1}}
\address[HE]{School of Science and Technology, Nottingham Trent University, Nottingham NG11 8NS, UK}
\ead{jun.he@ntu.ac.uk}
\cortext[cor1]{Corresponding author}

\tnotetext[ref1]{The first author was partly supported by the National Science Foundation of China (61303028) and the Fundamental Research Funds for the Central Universities (WUT: 2020IB006). The second author was supported by EPSRC under Grant No. EP/I009809/1.}

\tnotetext[ref2]{The paper is an extension of our conference paper: Chen, Y. and He, J. (2019). Average Convergence Rate of Evolutionary Algorithms II. In \textit{Int. Symposium on Intelligence Computation and Applications}. Springer.}

\begin{abstract}
The average convergence rate (ACR) measures how fast the approximation error of an evolutionary algorithm converges to zero per generation. It is defined as the geometric average of the reduction rate  of the approximation error over consecutive generations. This paper makes a theoretical analysis of the ACR in continuous optimization.   The obtained results are summarized as follows.  According to the limit property,   the ACR is classified into two categories: (1) linear ACR whose limit inferior value is larger than a positive  and (2) sublinear ACR whose value  converges to zero. Then, it is proven that the  ACR  is linear for evolutionary programming using  positive landscape-adaptive mutation, but sublinear for that using landscape-invariant or zero landscape-adaptive mutation.     
The relationship  between the ACR  and  the decision space dimension is also classified into two categories: (1) polynomial ACR whose value is larger than the  reciprocal of a polynomial function of the dimension for any generation, and (2) exponential ACR whose value is less than the  reciprocal of an exponential function of the dimension for an exponential long period. It is proven that for easy problems such as  linear functions, the ACR of the (1+1) adaptive  random univariate search  is polynomial. But for hard functions such as the deceptive function,  the ACR of both the (1+1) adaptive  random univariate search and evolutionary programming  is exponential.
\end{abstract}

%
\begin{highlights}
\item Average convergence rate measures the convergence speed of evolutionary algorithms by the geometric average of the reduction rate of the approximation error over consecutive generations.
\item   Evolutionary programming with landscape-invariant or zero landscape-adaptive mutation converges slowly, while that with positive landscape-adaptive mutation converges fast.
\item Average convergence rate are classified into two types:  polynomial or exponential as a function of the decision space dimension.
\end{highlights}

\end{frontmatter}


\section{Introduction}
In both empirical and theoretical studies of evolutionary algorithms (EAs), a fundamental question is how fast does an EA converge to the optimal solution of an optimization problem? In discrete optimization, this is usually measured by computational time, by either the hitting time~(the number of generations) or running time~(the number of fitness evaluations) when an EA first finds an optimal solution~\cite{he2001drift}. In continuous optimization, however, computational time often is infinite because  optimal solutions to many optimization problems are several points in a real coordinate space. Therefore, computational time has to be modified to the time when EAs reach an $\epsilon$-neighbor around  optimal  solutions~\cite{chen2011drift,agapie2013convergence,huang2020experimental}.  

The convergence rate  is an alternative way to evaluate the performance of EAs in continuous optimization. It quantifies how fast an EA converges   to the optimal solution set per generation in the decision space. So far, numerous theoretical work has been reported to discuss the convergence rate from different perspectives~\cite{rudolph1997local,rudolph1997convergence,he1999convergence,he2001conditions,auger2005convergence,auger2016linear,he2017initial}. 

Because of the equivalence of convergence in decision space and that in the objective space, it is  rational to investigate how fast the approximation error converges to zero in the objective space. 
The convergence rate discussed in this paper is defined in the objective space in terms of the  approximation error. Denote the fitness value of the best individual in the $t^{th}$ generation population as $f_t$  and  the approximation error as $e_t = | f_t-f^*|$ where $f^*$ represents the optimal fitness.  
The geometric convergence rate $e_t\le e_0 c^t$  can be obtained from the one-step convergence rate (CR), $e_t / e_{t-1}$, under the condition $e_t/e_{t-1}<c$~\cite{rudolph1997convergence,tarlowski2018geometric}. But unfortunately,  randomness  in EAs results in the oscillation of  $e_t/e_{t-1}$,   which in turn hinders its practical application in computer experiments. Instead, the geometric average of $e_t / e_{t-1}$ over consecutive $t$ generations is proposed as the average convergence rate (ACR)~\cite{he2016average} $ACR_t= 1-(e_t/e_0)^{1/t}$. A major advantage of $ACR_t$ is that it is more stable than $e_t/e_{t-1}$ in computer simulation. 

The ACR has been adopted as a practical metric of the convergence speed of EAs in continuous optimization~\cite{dong2018novel,dhivyaprabha2018synergistic,janiga2019self,li2019binary}. Although some theoretical results have been obtained for EAs in discrete optimization~\cite{he2016average}, there is no analysis of EAS in continuous optimization.  The current paper aims to extend the study from discrete optimization to  continuous optimization and  answer three research questions: When does the ACR converge to zero? When not? What is the relationship between the ACR and the decision space dimension?

The paper is organized as follows: Section~\ref{secWork} reviews the related work. Section~\ref{secEmpirical}  presents an empirical study of the ACR.   Section~\ref{secStudy} provides a general theoretical study of the ACR. Section~\ref{secEP} analyses the ACR of evolutionary programming. Section \ref{secRelation} investigates the relationship between the ACR and decision space dimension. Finally, Section~\ref{secConclusions} concludes the paper.

\section{Literature Review of Convergence Rate}
\label{secWork}
The   convergence rate of EAs  has been investigated from different perspectives and in varied terms. He, Kang   and Ding~\cite{he1999convergence,ding2001convergence}  studied the convergence in distribution by considering sequence $\{\parallel \mu_t-  \pi \parallel,t=1,2,\dots\}$ where $\mu_t$ is the probability distribution of the $t^{th}$ generation population $X_t$ and $\pi$ a stationary probability distribution.    Based on the Doeblin condition, they obtained an upper bound  $ (1-\delta)^{t-1}$ on $\parallel \mu_t-  \pi \parallel$ for some $\delta \in (0,1)$. He and Yu~\cite{he2001conditions} derived  lower and upper bounds on $1-\mu_t({X}^*_{\delta})$ where $\mu_t(\mathcal{S}^*_{\delta})$ denotes the probability of $X_t$ entering in a $\delta$-neighbor of  $X^*$ where $X^*$ denotes the set of optimal solutions.

Rudolph~\cite{rudolph1997local} compared  Gaussian and Cauchy mutation for minimization of the sphere function    in terms of the rate of local convergence, $\mathbb{E}[\min\{ \parallel X_{t+1} \parallel^2/ \parallel X_t \parallel^2,1\} \mid X_t]$, where $\parallel \cdot \parallel$ denotes the Euclidean norm. He proved that the rate is identical for Gaussian and spherical Cauchy distributions, whereas nonspherical Cauchy mutations lead to slower local convergence. Rudolph~\cite{rudolph1997convergence} also proved  under the condition $e_t/e_{t-1} \le c <1$, the sequence $\{e_t\}$ converges in mean geometrically fast to $0$, that is, $q^t e_t=o(1)$ for some $q>1$.  For a superset of the class of quadratic functions,   sharp bounds on the convergence rate are obtained.

Semenov and Terkel~\cite{semenov2003analysis} studied the convergence velocity of a simple EA with self-adaptation using a stochastic Lyapunov function and martingale theory. They proved that the velocity is asymptotically exponential $|x_t| \le \exp(-at)$ on the class of unimodal functions with the aid of Monte Carlo simulation. 

Beyer~\cite{beyer2013theory} developed a systematic theory of evolutionary strategies (ES) based on the progress rate and quality gain. The progress rate measures the distance change to  the optimal solution  in one generation,  $\mathbb{E}[\parallel X_t-X^* \parallel-\parallel X_{t-1}-X^* \parallel]$. The quality gain is  the fitness change in one generation,   $\mathbb{E}[ \bar{f}(X_t)-\bar{f}(X_{t-1}) ]$, where $\bar{f}(X)$ is the fitness mean of individuals in population $X$. Beyer et al.~\cite{beyer2014dynamics,beyer2016dynamics} analyzed dynamics of ES with cumulative step size adaption and ES with self-adaption and multi-recombination on the ellipsoid model and derived the quadratic progress rate. Akimoto et al.\cite{akimoto2017quality} investigated ES with weighted recombination
on general convex quadratic functions and derived
the asymptotic quality gain. However, Auger and Hansen~\cite{auger2006reconsidering} argued the limit  of the predictions using the progress rate.

Auger and Hansen~\cite{auger2011theory} developed the theory of ES from a new perspective using the stability of Markov chains. Auger~\cite{auger2005convergence} investigated the $(1,\lambda)$-SA-EA on the sphere function and proved the convergence of $(\ln \parallel X_t \parallel)/t$ based on Foster-Lyapunov drift conditions.
Jebalia et al.~\cite{jebalia2011log} investigated convergence rate of  the scale-invariant (1+1)-ES in minimizing the noisy sphere function and proved a log-linear convergence rate in the
sense that: $(\ln \parallel X_t \parallel)/t \to \gamma$ for some $\gamma$ as $t \to +\infty$.
Auger and Hansen~\cite{auger2016linear} further investigated the comparison-based step-size adaptive randomized search on scaling-invariant objective functions and proved  as $t \to +\infty$, $\ln (\parallel X_t \parallel/\parallel X_0 \parallel_t)/t \to -CR$ for some positive $CR$. This log-linear convergence can be regarded as an extension of the classical average rate of convergence in deterministic iterative methods~\cite{varga2009matrix}.

The above convergence rates are `evaluated by Markov chain analysis, however, process is complicated from theoretical and practical point of view'~\cite{janiga2019self}. Unlike them, the ACR has an attention to a close link with practice. 
Although optional to define a convergence rate in various spaces, it is preferred in this paper to investigate it in the objective space. With respect to many applications of EAs,  their performance  is often evaluated by the approximation error of obtained solutions. Thus, the convergence rate of EAs is defined as  the average reduction rate of the approximation error over consecutive generations in the objective space~\cite{he2016average}. For  discrete optimization, it has been proved~\cite{he2016average} that  under particular initialization, the ACR is equal to the spectral radius of a transition submatrix corresponding to transition probabilities within non-optimal solutions, and under random initialization (any solution can be chosen into the initial population with a positive probability), the ACR converges to the spectral radius. However,  no similar analysis has been done for the ACR in continuous optimization.

\section{Empirical Study of Average Convergence Rate}
\label{secEmpirical}
\subsection{Definition of Average Convergence Rate}
\label{subsecDefinition}
Consider a  minimization problem: 
\begin{equation}\label{equOP}
 \min f(\mathbf{x}), \quad {\mathbf{x}} =(x_1, \cdots, x_d)   \in  \mathcal{D} \subset \mathbb{R}^d,
\end{equation}
where $f(\mathbf{x}) $ is a continuous  function. $\mathcal{D}$ is the definition domain (called the decision space) and is bounded.  $d$ is the dimension. Denote the minimal function value as $f^*:=\min \{ f(\mathbf{x})\mid    {\mathbf{x}}  \in  \mathcal{D}\}$ and the optimal solution set as  $X^*: = \{\mathbf{x}  \in \mathcal{D} \mid f(\mathbf{x}) =f^* \}$.

Algorithm~\ref{alg1} describe a general framework of EAs. In EAs, an individual is a single point (solution) $\mathbf{x}$. A population $X$ is a union of finite individuals $X=(\mathbf{x}_1, \cdots, \mathbf{x}_{\mu})$ where $\mu$ is the population size. An optimal population $X$ satisfies $X \cap X^* \neq \emptyset$ and a non-optimal population $X$ satisfies $X \cap X^* = \emptyset$. Let $\mathcal{S}$ denote  the set   of all populations and $\mathcal{S}^*$  the set of optimal populations.

\begin{algorithm}[ht]
\caption{Evolutionary Algorithm}
\label{alg1}
\begin{algorithmic}[1]
\STATE generation counter  $t \leftarrow 0$;
\STATE $X_0 \leftarrow$ initialize a population  of individuals;
\WHILE{the stopping criterion is not satisfied}
\STATE $X_{t+1} \leftarrow$ generate a population of individuals from $X_{t}$ subject to a conditional transition probability $\Pr(X_{t+1} \mid X_0, \cdots, X_t)$;
\STATE  $t\leftarrow t+1$;
\ENDWHILE
\end{algorithmic}
\end{algorithm}

Given an initial population $X_0$,
the  \emph{fitness} of the population $X_t$ is  $f(X_t\mid X_0):=\min\{ f(\mathbf{x}) \mid \mathbf{x} \in X_t \}$, and its \emph{approximation error}  is $e(X_t \mid X_0):=|f(X_t)-f^*|$. Thereafter,  $f(X_t\mid X_0)$ and $e(X_t\mid X_0)$ are denoted by  $f(X_t )$ and  $e(X_t )$ in short respectively.
An  EA using \emph{elitist selection} always keeps the best found individual, that is, $e(X_{t+1})\le e(X_t)$.
Let   $f_t:=\mathbb{E}[{\mathbb{E}[ f(X_t)]\mid X_0}]$ and $e_t:=\mathbb{E}[\mathbb{E}[ e(X_t) \mid X_0]]$ denote the expected values of the fitness and approximation error respectively. In computer simulation, $f_t$ is calculated as the the average value over a number of runs.

An EA is called \emph{convergent in mean}~\cite{rudolph1997convergence} if starting from any $X_0$,  $\lim_{ t \to +\infty} e_t=0$. An EA is called  \emph{convergent almost surely}~\cite{rudolph1997convergence} if starting from any $X_0$, the probability $\Pr(\lim_{t\to +\infty} e(X_t)=0)=1.$
Given an approximation error sequence   $\{e_t; t=0,1, \cdots \}$, its \emph{one-step convergence rate} (CR) is the reduction rate of the approximation error at  at the $t^{th}$ generation.
\begin{align}
\label{equCR}
 CR_t :=
\frac{e_{t}}{e_{t-1}},  \quad t \in \mathbb{Z}^+,
\end{align}
where $\mathbb{Z}^+$ denotes the set of positive integers.
The \emph{average convergence rate for $t$ generations} (ACR)~\cite{he2016average} is defined by
\begin{align}
\label{equAverageRate}
   ACR_t :=
 1- \left(  \frac{e_t}{e_0} \right)^{1/t}= 1-\left(  \prod^{t}_{k=1} CR_k  \right)^{1/t}, \quad t \in \mathbb{Z}^+.
\end{align}
In (\ref{equAverageRate}), the term $(e_t/e_{0})^{1/t}$ represents a geometric average of the CR  over $t$ consecutive generations. $1-  (e_t/e_{0})^{1/t}$  normalizes the average between $(-\infty,1]$.  The ACR rate measures the convergence speed of an EA because the larger ACR, the faster convergence. A negative value of the ACR means the EA moves away from the optimal point. $ACR_t=1$ when  $e_t=0$ or equivalently the optimal solution is generated.

Similar to the convergence rate in deterministic iterative methods~\cite[Definition 3.1]{varga2009matrix},   another average  convergence rate of EAs is defined in the logarithmic form~\cite{he2016average}. 
\begin{equation}
\label{equLogRate}
ACR'_t:=-\frac{1}{t}\ln   \frac{  e_t }{ e_{0} }. 
\end{equation}
However, this rate cannot be adopted in computer simulation because its value becomes $+\infty$ when a population approximates an optimal solution, that is, $e_t = 0$.

\begin{figure}[ht]
\centering
\subfloat[][]{\includegraphics[height=4cm]{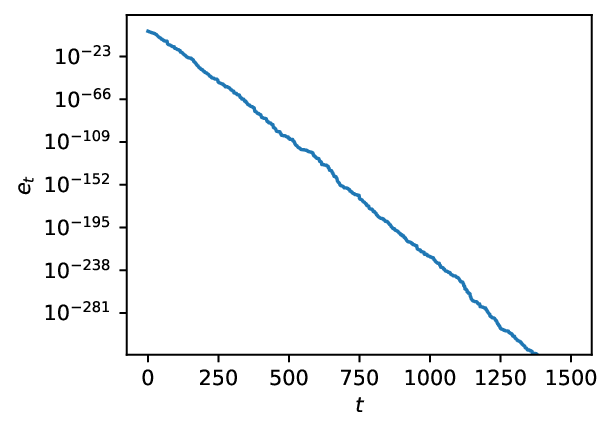}}%
\subfloat[][]{\includegraphics[height=4cm]{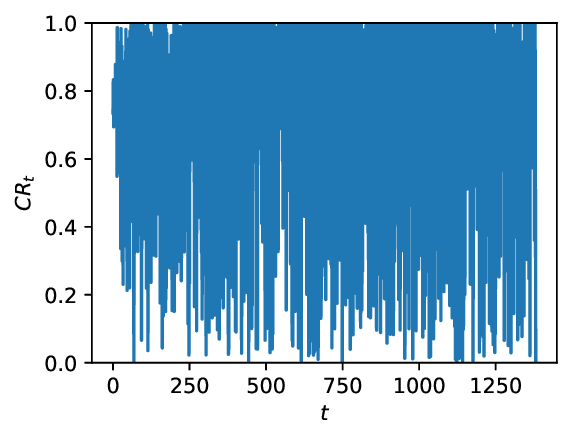}}%
\subfloat[][]{\includegraphics[height=4cm]{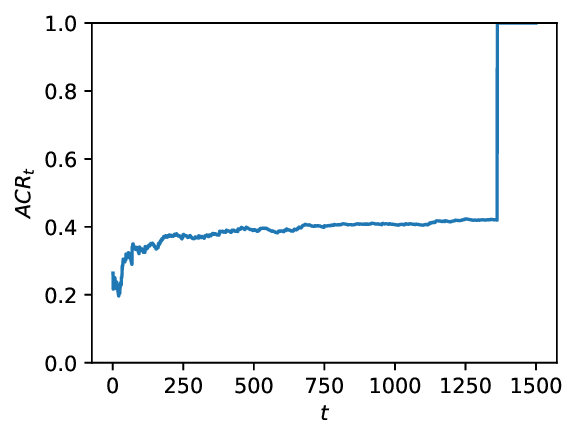}}
\caption{(a) $e_t$ converges to 0. (b) $CR_t$ oscillates between 0 and 1. (c) $ACR_t$ is stable at a constant 0.4.}%
\label{fig1}%
\end{figure}

Figure~\ref{fig1} compares the approximation error, CR and ACR  through an example. Figure~\ref{fig1}(a) shows $e_t$ decreases as $t$, but it does not quantify the convergence speed.  Figure~\ref{fig1}(b) shows $CR_t$ oscillates significantly as $t$.   Figure~\ref{fig1}(c) depicts that $ACR_t$ is more stable because it averages the CR values over consecutive generations. The ACR increases from 0.2 to 04, then jumps to 1 when the optimal solution is found.   The jump means that the EA has found the optimal solution. This phenomenon could happen in computer simulation because the number of runs is finite  and the EA may find the optimum in all runs.  

Figure~\ref{fig1}(b) and (c) reveals an essential difference  between the ACR and CR. The sequence $\{CR_t;  t \in \mathbb{Z}^+\}$   oscillates, but the sequence $\{ACR_t; t \in \mathbb{Z}^+\}$ is more stable. In practice, the calculation of $ACR_t$ is easy and it results in an exact erro expression $e(t)= e_0 (ACR_t)^t $. However, it is difficult to obtain the same expression from $CR_t$.

In the above ACR definition, the optimal fitness value $f^*$ is required. In case of $f^*$ unknown,  an \emph{alternative average convergence rate} is  defined by~\cite{he2016average}.
\begin{align}
\label{equAverageRate-alternative}
ACR^{\dagger}_t:=1-\left|\frac{f_{t+\tau}-f_t}{f_t -f_{t-\tau}}\right|^{1/\tau}, \quad\mbox{if } t \ge \tau,
\end{align}
where $\tau$ is an appropriate and user-defined time interval. In computer simulation, it is necessary to set $\tau$ to a large value. This reduces noise in calculating $ACR^{\dagger}_t$. It should be mentioned that the alternative ACR is not defined for $t <\tau$.

\subsection{Empirical Study of ACR and Alternative ACR}
\label{subsecEmpirical} 
Using the ACR or alternative ACR, it is convenient to quantify and visualize the convergence speed of EAs. Let us show this claim through computer simulation.

For the purpose of illustration, consider  (1+1) evolutionary programming  (Algorithm~\ref{alg2}) for minimizing the 2-d sphere function as an example. The sphere function is a unimodal function which is often used as a benchmark in EAs~\cite{qu2016novel}.
\begin{align}
\label{equSphere}
&\min f_S(\mathbf{x}) =x^2_1+x^2_2, \quad \mathbf{x} \in \mathbb{R}^2.
\end{align}
The minimal point to this function is $\mathbf{x}^*=(0,0)$ with $f^*=0$.

Evolutionary programming (EP) is a type of EAs which employs mutation and selection but without recombination.  (1+1) EP is equivalent to (1+1) evolutionary strategies (ES)  without crossover. However, a population-based ES employs a recommendation operator~\cite{back1993evolutionary}.
\begin{algorithm}[ht]
\caption{(1+1) Evolutionary Programming}
\label{alg2}
\begin{algorithmic}[1]
\STATE  generation counter $t \leftarrow 0$;
\STATE  initialize an individual  $\mathbf{x}_{0}$;
\WHILE{the maximal number of generations is not reached}
\STATE   generate a new individual  $\mathbf{y}_{t} = \mathbf{x}_{t}+\mathbf{z}_t$ where $\mathbf{z}_t$ obeys a probability distribution (such as Gaussian, Cauchy or L\'evy) distribution;
\STATE select the best one from $\mathbf{y}_t$ and $ \mathbf{x}_t $ as $\mathbf{x}_{t+1}$;
\STATE $t\leftarrow t+1$;
\ENDWHILE
\end{algorithmic}
\end{algorithm}

A child $\mathbf{y}$ is generated by Gaussian mutation $\mathbf{y}=\mathbf{x}+\mathbf{z}$,
where    $\mathbf{z}=(z_1, \cdots, z_d)$  obeys the Gaussian probability distribution $z_i\sim \mathcal{N}(0, \sigma_i)$.
Two variants of Gaussian mutation   are chosen with different settings of $\boldsymbol{\sigma}=(\sigma_1, \cdots, \sigma_d)$.

\begin{itemize}
\item \textbf{Invariant Gaussian mutation:} $\boldsymbol{\sigma}$ is set to  constants, that is, for any $i$, $\sigma_i$ is a constant. The (1+1) EP using invariant Gaussian mutation is called the \emph{(1+1) invariant EP}.
\item \textbf{Adaptive Gaussian mutation:}   $\boldsymbol{\sigma}$  varies as $\mathbf{x}$, that is, $\boldsymbol{\sigma}$ is a function of $\mathbf{x}$. The (1+1) EP using adaptive Gaussian mutation is called the \emph{(1+1) adaptive EP}.
\end{itemize}

In computer simulation, set $\sigma_i=1$ in the (1+1) {invariant} EP and $\sigma_i=|x|_i$  in the (1+1) {adaptive EP}.   The adaption is based on  commonsense in the design of EAs:  as $\mathbf{x}$ is close to the optimal solution $0$, $\boldsymbol \sigma$ is set to a small value. This is equivalent to the practical strategy:  as $t$ increases, $\mathbf{x}_t$ is close  to the optimal solution, then $\boldsymbol \sigma$ is reduced.  $\mathbf{x}_0$ is randomly generated in $[-20,20]^2$.  Each algorithm runs $1,000$ times independently and $f_t$ is the average over the 1,000 runs. The maximum number of generations is $300$. The time interval $\tau$ for calculating the alternative ACR is chosen to be $50$.

\begin{figure}[ht]
\centering
\subfloat[][]{\includegraphics[height=4cm]{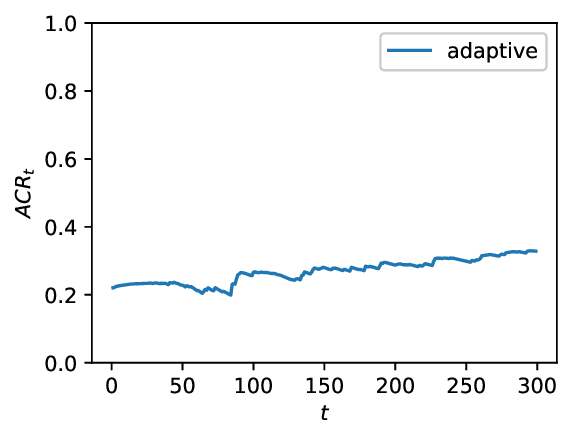}}%
\subfloat[][]{\includegraphics[height=4cm]{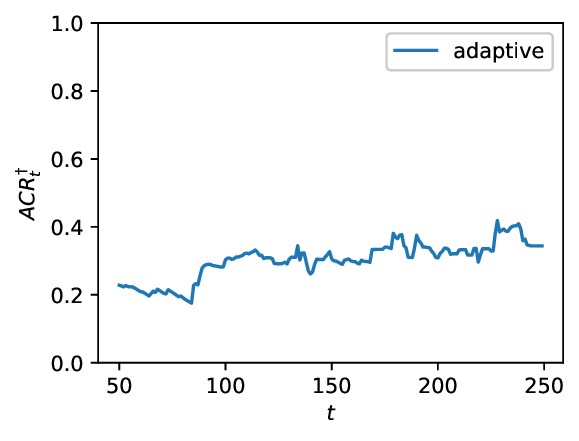}}
\caption{(a) $ACR_t$  of the (1+1) adaptive EP on the sphere function.(b)  $ACR^{\dagger}_t$  }%
\label{figAdaptive}%
\end{figure}

The first experiment is to compare the ACR and alternative ACR of the adaptive  EP on the sphere function. Trend plots of the ACR and alternative ACR  are illustrated in Figure~\ref{figAdaptive}.  Figure~\ref{figAdaptive}(a) shows that the ACR of the adaptive EP  increases to a positive, while  
Figure~\ref{figAdaptive}(b) depicts similar tends of the alternative ACR.  But $ACR^{\dagger}_t$ suffers a little bigger noise than $ACR_t$ due to the introduction of the $\tau$-th order difference.

\begin{figure}[ht]%
\centering
\subfloat[][]{\includegraphics[height=4cm]{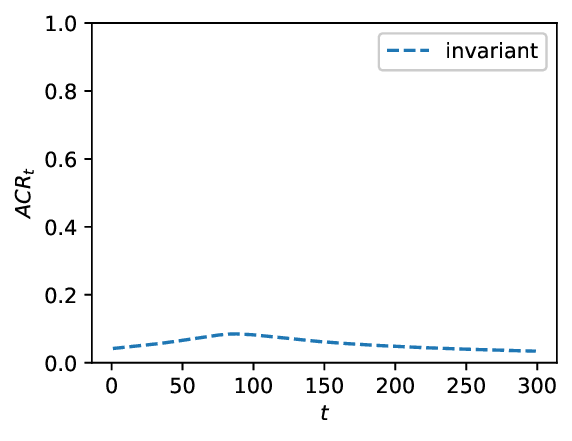}}%
\subfloat[][]{\includegraphics[height=4cm]{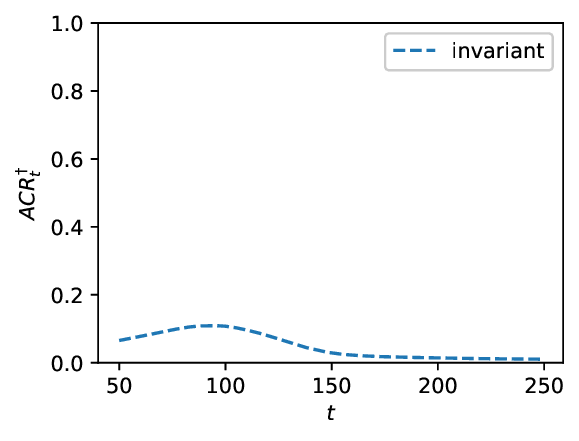}}
\caption{(a) $ACR_t$ of the (1+1) invariant EP on the sphere function. (b) alternative $ACR^{\dagger}_t$.}%
\label{figInvariant}%
\end{figure}

The second experiment is to compare  the ACR and alternative ACR of the invariant EP on the sphere function. Trend plots of the ACR and alternative ACR  are illustrated in Figure~\ref{figInvariant}.  Figure~\ref{figInvariant}(a) shows that the ACR decreases as time, while  
Figure~\ref{figInvariant}(b) depicts a similar tend for the alternative ACR.

The results reveal that the adaptive EP  converges faster than the invariant EP. For the adaptive EP, its ACR and alternative ACR both tend to a positive constant between 0.38 to 0.4. In general, an ACR is called \textit{linear} if it tends towards a positive.  But for the invariant EP, its ACR and alternative ACR decreases to a smaller constant and eventually towards 0. In general, an ACR is called \textit{sublinear} if it converges to 0.  

\subsection{Comparison between ACR and CR}
\label{subsecComparison}
The ACR is more stable than the CR in numerical calculation. Let us show the claim through computer simulation. 

\begin{figure}[htb]%
\centering
\subfloat[][]{\includegraphics[height=4cm]{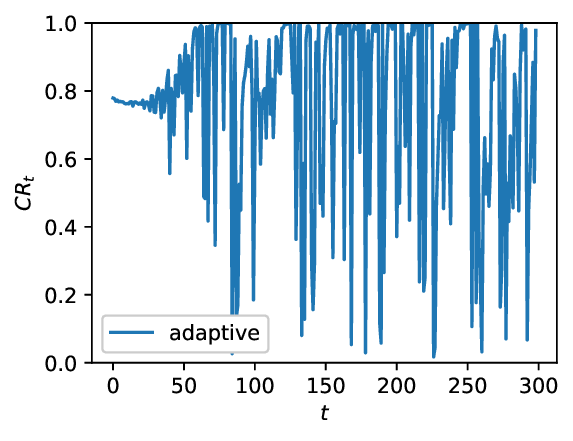}}%
\subfloat[][]{\includegraphics[height=4cm]{f1-ACR-ada.eps}}%
\caption{(a) CR of (1+1)  adaptive  EP  oscillates significantly  on the sphere function. (b)  ACR is more stable.}%
\label{figComparisonCRandACR}%
\end{figure}

The first experiment is to compare the ACR and CR of the (1+1) adaptive EP on the sphere function.
Experimental setting is the same as that in the previous subsection.   Figure~\ref{figComparisonCRandACR} illustrates trend plots of the ACR and CR.  The CR fluctuates greatly between 0 and 1. It is impossible to quantify the convergence speed using the CR. But the ACR clearly converges to a positive constant around 0.38.

\begin{figure}[htb]%
\centering
\subfloat[][]{\includegraphics[height=4cm]{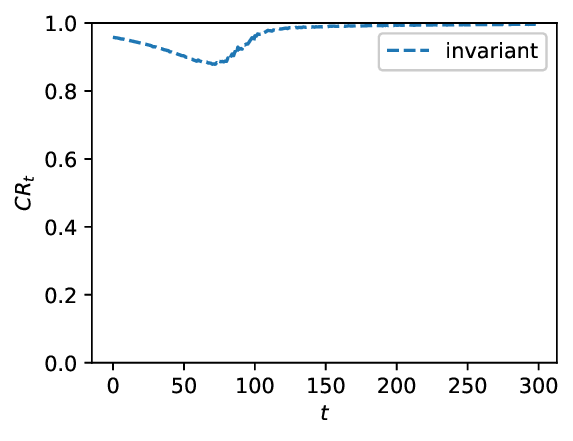}}%
\subfloat[][]{\includegraphics[height=4cm]{f1-ACR-inv.eps}}%
\caption{(a)  CR   of the (1+1) invariant EP on the sphere function. (b)  ACR.}%
\label{figLinearRate2}%
\end{figure}

The second experiment is to compare the ACR and CR of the (1+1) invariant EP on the sphere function.
Experimental setting is the same as that in the previous subsection.   Figure~\ref{figLinearRate2} illustrates trend plots of the ACR  and CR. The figure shows that the CR converges to 1 and the ACR to 0. The difference is caused by the normalization in the ACR to ensure the lower ACR, the slower convergence. Figure~\ref{figLinearRate2}(b) reveals that the convergence speed of the (1+1) invariant EP eventually decreases to 0. 

The experimental results confirm  that the numerical calculation of the ACR   is more stable than the CR.

\subsection{Relationship between ACR and Decision Space Dimension}
\label{subsecNumRelation}
The ACR of  EAs will decrease as the decision space dimension increases. Let us verify this claim through computer simulation.

For the sake of illustration, consider a (1+1) random univariate search method (RUS), described in Algorithm~\ref{alg3}, as an example. This algorithm can be regarded as  random local search in continuous optimization because both make a random one-dimensional search at each generation. The RUS using adaptive Gaussian mutation  is called the \emph{adaptive RUS} in short.

\begin{algorithm}[ht]
\caption{(1+1) Random Univariate Search}
\label{alg3}
\begin{algorithmic}[1]
\STATE  $t \leftarrow 0$;
\STATE  initialize a solution $\mathbf{x}_{0} =(x_1, \cdots, x_d) $ ;
\WHILE{the maximal number of generations is not reached}
\STATE   choose one index $j \in \{1, \cdots, d\}$ at random, and generate a new solution by $\mathbf{y}_t= \mathbf{x}_{t}+\mathbf{z}_t$  where $\mathbf{z}_t=(z_1, \cdots, z_d)$,  $z_{j}\sim \mathcal{N}(0,\sigma_j)$ is a Gaussian random variable and $z_{i}=0$ for other $i\neq j$; if $\mathbf{y}_t$ is out of the definition domain, let $\mathbf{y}_t=\mathbf{x}_t$;
\STATE  select the best one from $\mathbf{y}_t$ and $ \mathbf{x}_t $ as $\mathbf{x}_{t+1}$;
\STATE $t\leftarrow t+1$;
\ENDWHILE
\end{algorithmic}
\end{algorithm}

Two test functions are used in computer simulation.
The functions are inspired from  the OneMax function and deceptive function in pseudo-Boolean function  optimization. The OneMax function is  the easiest to a (1+1) elitist EA and the deceptive function is  the hardest~\cite{he2015easiest}. A variant  OneMax function in continuous optimization is defined by  
\begin{align}
\label{equOneMax}
 \max f_O(\mathbf{x}) :=d-\textstyle \sum^d_{i=1} x_i, \quad \mathbf{x} \in [0,1]^d.
\end{align}where $\mathbf{x}^*=(0, \cdots, 0)$ and $f^*=d$.

A deceptive function in continuous optimization is defined by 
\begin{equation}\label{equdeceptive}
 \max f_D(\mathbf{x})=\left\{
\begin{array}{lll} \textstyle \sum_{i=1}^d x_i,& \textstyle \mbox{ if }\sum_{i=1}^d x_i \ge 1/2,\\
\textstyle d+1-\sum_{i=1}^d x_i, &\textstyle  \mbox{ if } \sum_{i=1}^d x_i<1/2,
\end{array}
\quad \mathbf{x} \in [0,1]^d.
\right.
\end{equation}
The global optimum is $\mathbf{x}^*=(0, \cdots, 0)$ and $f^*=d+1$. The basin of attraction of $\mathbf{x}^*$ is $\{\mathbf{x} \mid \sum_{i=1}^d x_i\le 1/2\}$.
The  deceptive function (\ref{equdeceptive}) has a local optimum at $(1,\cdots,1)$.

In computer simulation, set $\sigma_j=x_j$ for the selected index $j$ in the  adaptive RUS.   $\mathbf{x}_0$ is chosen  from $[0,1]^d$ at random. The algorithm runs $2,000$ times independently on each test function. $f_t$ is the average over the 2,000 run.  The maximum number of generations is  $500$.

\begin{figure}[ht]%
\centering
\subfloat[][]{\includegraphics[height=4cm]{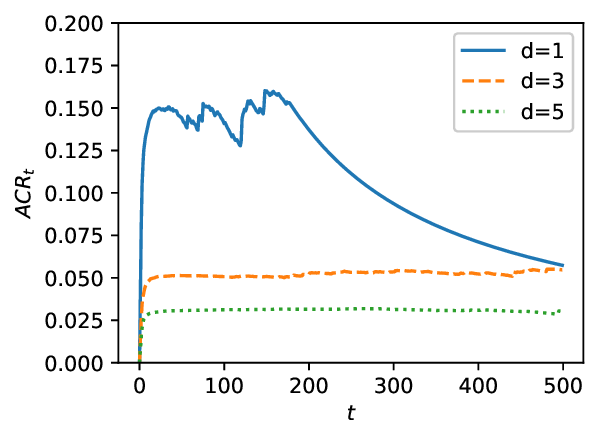}}%
\subfloat[][]{\includegraphics[height=4cm]{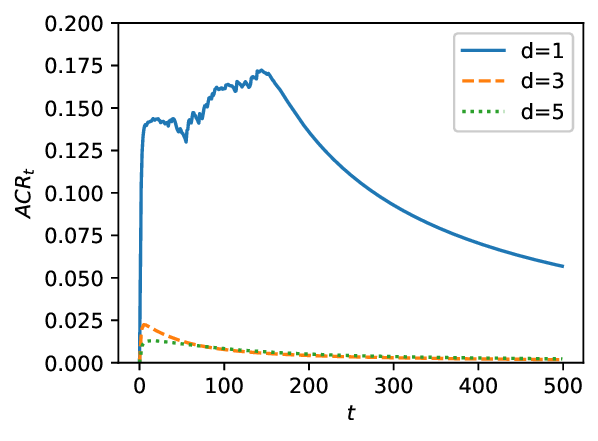}}%
\caption{(a) The ACR of adaptive RUS  decays slowly as   dimension $d=1,3, 5$  on the variant OneMax function. (b) But it decays quickly as dimension $d=1,3, 5$  on the deceptive function.}%
\label{figOneMax}%
\end{figure}

The experiments is to compare the ACR between the variant OneMax function and deceptive function. Trend plots of the ACR  are  illustrated in Figure~\ref{figOneMax}.  Figure~\ref{figOneMax}(a) shows that  the ACR  on the variant OneMax function converges to some positive constants over generations for $d=1,3,5$. But Figure~\ref{figOneMax}(b) depicts  the ACR on the deceptive function  decreases quickly as the dimension $d$ increases.

The experimental results demonstrate that the ACR decreases as the dimension increases. For an easy function like the OneMax function, the ACR decreases slowly as the dimension.   In theory, it is expected that the ACR is larger than the  reciprocal of a polynomial function of $d$.   But for a hard function like the deceptive function, the ACR decreases quickly as the dimension increases.   In theory, it is expected that the ACR is less than the  reciprocal of an exponential function of $d$. 

\section{General Theoretical Study of Average Convergence Rate}
\label{secStudy}
\subsection{Advantage of ACR over CR}
\label{secNecssity}  
Computer simulation in Section~\ref{subsecComparison} shows an advantage of the ACR over the CR, that is, when a CR sequence oscillates, the ACR sequence is still stable. This subsection explains this difference.  

The terms of linear, sub-linear or super-linear convergence has been used  in describing the convergence speed of an iterative sequence. In EAs, a sequence $\{CR_t, t \in \mathbb{Z}^+\}$ converges \emph{linearly} if $\lim_{t \to +\infty} CR_t  <1$ or converges \emph{sublinearly} if  $\lim_{t \to +\infty} CR_t =  1$.
Similarly, a  sequence $\{ACR_t, t \in \mathbb{Z}^+ \}$  converges \emph{linearly} if $\lim_{t \to +\infty} ACR_t >0$  or converges \emph{sublinearly}  if  $\lim_{t \to +\infty} ACR_t =  0$.

From the definition
\begin{align}
    ACR_t =  1-\left(\prod^{t}_{k=1} CR_k \right)^{1/t},
\end{align}
the linear   convergence of the ACR can be derived from the linear convergence of the CR. But the inverse does not hold. 
This is the main advantage of the ACR over the CR. Let us verify the claim using an example.

Let us consider an example which is  (1+1) EP  (Algorithm~\ref{alg-EP})  combining Cauchy and Gaussian mutation together. Because Cauchy and Gaussian mutation operators have different probability density functions, combination of them could result in faster convergence~\cite{chellapilla1998combining,yao1999evolutionary,dong2007evolutionary,mallipeddi2010ensemble}. 

\begin{algorithm}[ht]
\caption{(1+1) Evolutionary Programming}
\label{alg-EP}
\begin{algorithmic}[1]
\STATE  generation counter $t \leftarrow 0$;
\STATE  initialize an individual  $\mathbf{x}_{0}$;
\WHILE{the maximal number of generations is not reached}
\STATE   generate a new individual by  Gaussian mutation  $\mathbf{y}_{t} = \mathbf{x}_{t}+\mathbf{z}_t$ where $\mathbf{z}_t$ obeys a probability distribution; if $\mathbf{y}_t$ is beyond the definition domain $\mathcal{D}$, let $\mathbf{y}_t=\mathbf{x}_t$;
\STATE select the best one from $\mathbf{y}_t$ and $ \mathbf{x}_t $ as $\mathbf{x}_{t+1}$;
\STATE $t\leftarrow t+1$;
\ENDWHILE
\end{algorithmic}
\end{algorithm}

Two mutation operators are alternately used in this (1+1) EP, that is,  Cauchy mutation is applied when $t$ is an odd number and  Gaussian mutation is applied when $t$ is an even number. Thus, $z_t$  obeys  Cauchy or Gaussian probability distribution, probability density functions of which are $p_c=  \frac{1}{\pi(1+x^2)}$ and $p_g=  \frac{1}{\sqrt{2\pi}}e^{-\frac{x^2}{2}}$, respectively.

The (1+1) EP is used to minimize a JUMP function which is a typical multi-modal problem. This  function is similar to the JUMP function in pseudo-Boolean optimization~\cite{jansen2014performance}. The the optimal solution set is $\{x; |x|<1\}$.
\begin{align}
\label{equJUMP}
&f_J(x)=\left\{
\begin{array}{lll}
    0  & \mbox{if }  |x| <1,\\
    4 -|x| & \mbox{if } 1 \le |x|<2,\\
    |x| &\mbox{otherwise.}
\end{array}\right.
\end{align}

For the sake of analysis, assume that the initial point ${x}_0=2$. After $t(t\ge 1)$ iterations, ${x}_t$ either jump to the flat $\{|x|<1\}$ or stay at $|x|=2$. Thus, the reduction rate of expected error is (without loss of generality, let $x_t=2$ or $|x_t| <1$)
\begin{align}\label{CR}
  &\frac{e_{t+1}}{e_t} 
  =\frac{\mathbb E[e({x}_{t+1})| {x}_t=2]\Pr(x_t=2)+0\cdot\Pr(|\mathbf{x}_t|<1)}{2\cdot\Pr({x}_t=2)+0\cdot\Pr(|{x}_t|<1)} 
  =\frac{1}{2}\mathbb E[e( {x}_{t+1})|\mathbf{x}_t=2].
\end{align}
Moreover, the conditional expectation of error change is for Cauchy mutation,
\begin{align}\label{IM_C}
  \mathbb E[e({x}_{t})-e({x}_{t+1})| {x}_t=2]=&\int_{-1}^{1}(2-0)\frac{1}{\pi}\frac{1}{1+(x-2)^2}dx=\frac{2}{\pi}\left(\arctan 3-\frac{\pi}{4}\right),
\end{align}
and for Gaussian mutation, 
\begin{align}\label{IM_G}
  \mathbb E[e({x}_{t})-e(\mathbf{x}_{t+1})| {x}_t=2]=&\int_{-1}^{1}(2-0)\frac{1}{\sqrt{2\pi}}e^{-\frac{x^2}{2}}dx=4\left(\Phi(1)-\frac{1}{2}\right), \end{align}
where $\Phi(\cdot)$ is cumulative distribution function (CDF) of the standard Gaussian distribution. From (\ref{CR}), (\ref{IM_C}) and (\ref{IM_G}) we know that for any integer $k \ge 1$
\begin{align*}
  CR_{2k}= \frac{e_{2k}}{e_{2k-1}}=C_a:=2\left(\Phi(1)-\frac{1}{2}\right),
\end{align*}
and
\begin{align*}
  CR_{2k+1}= \frac{e_{2k+1}}{e_{2k}}=C_b:=\frac{3}{4}-\frac{1}{\pi}\arctan 3.
\end{align*} Since $C_a \neq C_b$, the sequence $\{CR_t\}$ oscillates and does not converge.

However, from
\begin{align*}
ACR_{t} &=\left\{\begin{aligned}& 1-(C_a C_b)^{1/2},&& \mbox{if } t=2k, \\ & 1-(C_aC_b)^{1/2} (C_b/C_a)^{1/2t}, && \mbox{if } t=2k+1, \end{aligned}\right. k=1,2,\dots
\end{align*}
we get $ACR_t   \to 1-(C_aC_b)^{1/2}$ as $t \to +\infty$, that is, the sequence $\{ACR_t \}$ is convergent. 

This example shows that the CR sequence could oscillate and not converge  in some adaptive EAs.  However, the ACR sequence converges thanks to the average of the CR  for consecutive generations.

\subsection{Assumptions in the Theoretical Study}
\label{subsecAssumptions}
In order to make a theoretical analysis, EAs under investigation are assumed to satisfy several conditions. 
\begin{enumerate}
\item   (Supermartingale). The expected approximation error does not increase. For any non-optimal $X_0$ and any $t$,
\begin{equation}
\label{conSupermartingale}\mathbb{E}[e(X_{t+1}) \mid X_0, \cdots, X_t] \le   e(X_{t}).
\end{equation}  The sequence $\{e_t; t=0,1,\cdots\}$ is a monotonically decreasing function of $t$. This condition is different from elitism in EAs which requires $e(X_{t+1}) \le   e(X_{t})$.
A direct consequence from this condition is  $ACR_t \in [0,1]$.

\item    (Markov chain). 
 The state of $X_{t+1}$  depends on $X_t$ only. For any $X_0$ and any $t$, the transition probability
\begin{equation}
\label{conMarkov}
\Pr(X_{t+1} \mid X_0, \cdots, X_t )=\Pr(X_{t+1} \mid X_t ).
\end{equation}

\item   (Stochastic algorithm). Starting from any non-optimal $X_0$,  for any $t$, it holds 
\begin{equation}
\label{conStochastic}
\Pr(X_t \cap X^*)<1, \quad (\mbox{then } e_t >0).
\end{equation} 
If $\Pr(X_t \cap X^*)=1$,   an EA reaches the optimal set at the $t^{th}$ generation with probability 1. It degenerates to a deterministic-like algorithm.

\item  (Normal reduction). The reduction rate of the approximation error satisfies  
\begin{align}\label{conNormal}
    \lim_{t \to +\infty} \left(\frac{e_{t}}{e_{t-1}}\right)^{1/t}=1.
\end{align}
The condition holds for normal EAs. 
If this condition doesn't hold, $\lim_{t \to +\infty} (e_{t}/e_{t-1})^{1/t}<c<1$, then  for a large  $t$, $e_{t}/e_{t-1} <  c^t$. For example, let $c=0.9$ and $t=1000$, we have $e_{1000} <  1.75 \times 10^{-46} e_{999} $.  This rapid reduction rate is unpractical in normal EAs. 
\end{enumerate}
  Many EAs satisfy the above four conditions. By default, they are always assumed to be true in the theoretical  study in this paper. 

Markov chains associated with EAs can be classified into homogeneous and inhomogeneous, depending on whether  genetic   operators (mutation, crossover and selection)   change  over time~\cite{he2001conditions}.   This paper focuses on two types of EAs which are modeled by homogeneous Markov chains as below. 
\begin{enumerate}
    \item   The population sequence $\{X_t; t=0,1,\cdots\}$ is a \emph{homogeneous Markov chain}, that is, for any $t$, any $X$ and any subset $\mathcal{A} \subset \mathcal{S}$, the transition probability
\begin{equation}
\label{Chain1}
\Pr(X_{t+1} \in \mathcal{A} \mid   X_{t}=X)=P(X; \mathcal{A}).
\end{equation}
Transition probabilities do not change over time.
\item   A population subsequence $\{X_{\kappa t}, t=0,1, \cdots\}$ (where $\kappa >1$) is a homogeneous Markov chain.   Transition probabilities follow  a periodic change over generations. Let $Y_t =X_{\kappa t}$.
\begin{align}
\label{Chain2}
&\Pr(Y_{t+1} \in \mathcal{A} \mid   Y_{t}=X)=P^{(\kappa)}(X;  \mathcal{A}), \\
&\mbox{but } \Pr(X_{t+1} \in \mathcal{A} \mid   X_{t}=X) \neq \Pr(X_{t+2} \in \mathcal{A} \mid   X_{t+1}=X).
&\end{align}
The original population sequence $\{X_t; t=0,1,\cdots \}$ is an {inhomogeneous Markov chain}.  
\end{enumerate}

The current paper will not discuss other types of EAs whose  genetic   operators  change  over time. Their analysis needs further understanding of inhomogeneous genetic operators and advanced tools from inhomogeneous Markov chains or stochastic processes.    This topic is left for future research.

\subsection{Linear and Sublinear ACR}
\label{subsecClassificcation}
\begin{figure}[ht]
    \centering
    \includegraphics[height=40mm]{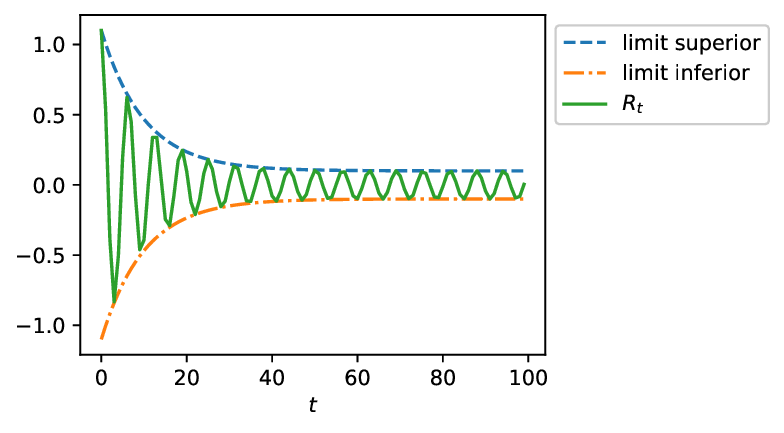}
    \caption{The limit superior and inferior of the sequence $R_t= \cos(t)  (0.1+\exp(-0.1 t))$.}
    \label{figLimit}
\end{figure}

Computer simulation in Section \ref{subsecEmpirical} demonstrates different trend plots of the ACR for the (1+1) invariant EP and (1+1) adaptive EP. The ACR sequence   has different limit properties.
In practice, it is required that $e_t$ converges to $0$, however this does not imply that $ACR_t$ will converge as $t\to \infty$. Thus,  the \emph{limit superior} and \emph{limit inferior}~\cite{snow2003exploratory} of the sequence $\{ACR_t \}$, defined as (\ref{LimSup}) and (\ref{LimInf}) respectively,  are introduced to describe the limit property.
\begin{align}
        \varlimsup_{t \to +\infty}  ACR_t :=\inf _{t \geq 0}\{ \sup_{s\geq t} \{ACR_{s}: s \geq t\}: t \geq 0\}.\label{LimSup}\\
      \varliminf_{t \to +\infty} ACR_t :=\sup _{t \geq 0}\{ \inf_{s\geq t} \{ACR_{s}: s \geq t\}: t \geq 0\},\label{LimInf}
\end{align}
where  $\inf$ is the abbreviation of  mathematical infimum and $\sup$  the abbreviation of  supremum.
An example of the  {limit superior and inferior}  is  illustrated in Figure~\ref{figLimit}. Consider the sequence $R_t= 0.1\cos(t)\times (\exp(-0.1t)+0.1)$ where $t=0,1,2, \cdots$. Although the sequence does not converge, its limit superior ($-0.1$) and inferior ($-0.1$) still exist.

Existing theoretical results~\cite{teytaud2006general,beyer2014convergence,auger2016linear} show that the convergence rate of EAs in continuous optimization is up to linear. 
Similarly, the ACR of EAs can be classified into two categories via its limit property.
\begin{itemize}
    \item The ACR is \emph{linear}, if its limit  inferior is a positive, that is, $\varliminf_{t \to +\infty}  ACR_t= C>0$. In this case, the approximation error  reduces geometrically fast to 0. For example, the ACR in Figure~\ref{figAdaptive} is linear.

    \item The ACR   is \emph{sublinear},  if it  asymptotically reduces  to zero as $t \to +\infty$, that is, $\lim_{t \to +\infty}  ACR_t = 0$. In this case  the approximation error  converges slowly to 0.  For example, the ACR in Figure~\ref{figInvariant} is sublinear. 
\end{itemize}

The ACR can be estimated using one-step transition or multi-step probability transition.   Given an error sequence $\{e_t, t=0,1, \cdots\}$,
the \emph{one-step error change  at the $t^{th}$ generation} is
$
    \Delta  e_t := e_t- e_{t+1}.
$ It is similar to the quality gain~\cite{beyer2013theory} but the latter is defined on  the fitness mean of individuals in a population $\bar{f}_t$ rather than the error mean $e_t$. The \emph{rate of one-step error change} is
$ {  \Delta  e_t}/{e_t}
$.  Given a positive integer $\kappa$,
the \emph{$\kappa$-step error change  at the $t^{th}$ generation} is
$
    \Delta^{(\kappa)}  e_t := e_t- e_{t+\kappa}.
$
 The \emph{rate of $\kappa$-step  error change} is
$ {  \Delta^{(\kappa)} e_t}/{e_t}
$.

\begin{theorem}
\label{theLimit}
(1)  For the error sequence $\{e_t; t=0,1,\cdots\}$,  if there exist a positive integer $\kappa$ and $0<C<1$,  the rate of {$\kappa$-step error change}  satisfies
\begin{equation}\label{equErrorChange}
  \varliminf_{ t \to +\infty } \frac{\Delta^{(\kappa)} e_{t}}{e_{t}} =1- C>0,
\end{equation}
 then the ACR is linear, that is,
$\lim_{ t \to +\infty } ACR_t =1-C^{1/\kappa}>0.$

(2) For the error sequence $\{e_t; t=0,1,\cdots\}$,  if
\begin{equation}\label{equErrorChange1} \lim_{ t \to +\infty }\frac{\Delta  e_t}{e_t} =0,
\end{equation}
then the ACR is sublinear, that is,
$\lim_{ t \to +\infty } ACR_t =0.$
\end{theorem}

\begin{proof}
(1)   While the rate of  $\kappa$-step error change   converges to a positive value, there are two different cases to be discussed for the number of iteration $t$.
\begin{enumerate}
  \item If $t=m\kappa$ for an integer $m>0$, we have
  \begin{align}\label{Case1}
    ACR_{m\kappa} =1-\left(\prod^{m\kappa}_{k=1} \frac{e_{k}}{e_k-1}\right)^{1/m\kappa}=1-\left[\prod^{m}_{l=1} \left(1-\frac{\Delta^{(\kappa)}e_{(l-1)\kappa}}{e_{(l-1)\kappa}}\right)\right]^{1/(m\kappa)}.
  \end{align}
  Note that $m$ tends to $+\infty$ when $t\to +\infty$. Then, from (\ref{equErrorChange}) and (\ref{Case1}) we know \begin{align}\label{Case11}\lim_{m\to +\infty}ACR_{m\kappa}=1-C^{1/\kappa}.\end{align}
  
  \item If $t=m\kappa+k$ for integers $m, k$ such that $m>0$, $0<k<\kappa$, we know
  \begin{align}\label{Case2}
    ACR_{m\kappa+k} &=1-\left(\prod^{m\kappa+k}_{k=1} \frac{e_{k}}{e_{k-1}}\right)^{1/(m\kappa+k)}=1-\left[\prod^{m}_{l=1} \left(1-\frac{\Delta^{(\kappa)}e_{(l-1)\kappa}}{e_{(l-1)\kappa}}\right)\left(\frac{e_{m\kappa+k}}{e_{m\kappa}}\right)\right]^{1/(m\kappa+k)}
  \end{align}
  From (\ref{equErrorChange}) we know that $\lim_{t\to +\infty}\frac{e_{t+\kappa}}{e_{t}}=C$, which implies that there exists $t_0>0$ such that
\begin{equation*}
  \frac{e_{t+\kappa}}{e_{t}}>\frac{C}{2},\quad\forall\,t>t_0.
\end{equation*}
Then, monotonicity of $e_t$ says that $\exists m_0>0$,
\begin{equation*}
  \frac{C}{2}<\frac{e_{m\kappa+\kappa}}{e_{m\kappa}}\le \frac{e_{m\kappa+k}}{e_{m\kappa}}\le 1,\quad\forall\,m>m_0,
\end{equation*}
and we conclude that
\begin{align}\label{term}
  \lim_{m\to +\infty}\left(\frac{e_{m\kappa+k}}{e_{m\kappa}}\right)^{1/(m\kappa+k)}=1,\quad\forall\,\,0<k<\kappa.
\end{align}
Combining (\ref{equErrorChange}), (\ref{Case2}) and (\ref{term}) we know
\begin{align}\label{Case22}
  \varliminf_{m\to +\infty}ACR_{m\kappa+k}=\varliminf_{m\to +\infty}\left\{1-\left[\prod^{m}_{l=1} \left(1-\frac{\Delta^{(\kappa)}e_{(l-1)\kappa}}{e_{(l-1)\kappa}}\right)\right]^{1/(m\kappa+k)}\right\}=1-C^{1/\kappa}.
\end{align} 
\end{enumerate}

From (\ref{Case11}) and (\ref{Case22}), we conclude that $\lim_{t\to +\infty}ACR_t=1-C^{1/\kappa}$.

(2) When $\lim_{ t \to +\infty }{\Delta  e_t}/{e_t} =0$, from the ACR definition, we know
$\lim_{t\to\infty}ACR_t=0$.
\end{proof}
To derive the linear ACR sequence, Condition~(\ref{equErrorChange}) requires   the rate of $\kappa$-step error change larger than a positive. This condition is weaker than the rate of the one-step error change larger than zero. But to  derive the sublinear convergence of the ACR sequence, Condition~(\ref{equErrorChange1}) requires  the rate of one-step error change to converge to 0.

\subsection{Link between ACR and Alternative ACR}
\label{subsectLink}
Computation simulation in Section~\ref{subsecEmpirical}   shows  the ACR and  alternative ACR have similar behaviors. There is a link between the ACR and alternative ACR. Under some conditions, their limits are identical.

\begin{theorem}
\label{thesimilarity}
For the error sequence $\{e_t\}$, if  for a positive integer $\kappa$ and $0<C<1$, 
\begin{equation}\label{equErrorChange2}
  \lim_{ t \to +\infty } \frac{\Delta^{(\kappa)} e_{t}}{e_{t}} =1- C>0,
\end{equation}
 and in the definition of the alternative ACR, choose $\tau$ to a multiple of $\kappa$, $$\lim_{t \to +\infty} ACR_t =\lim_{t \to +\infty} ACR^{\dagger}_t.$$
\end{theorem}
\begin{proof}
Since $\tau$ is a multiple of $\kappa$, we have $\tau=m\kappa$ for a positive integer $m$. Then,
\begin{align*}
  \lim_{t\to +\infty}ACR^{\dagger}_t & =1-\lim_{t\to +\infty}\left(\frac{e_{t}-e_{t+\tau}}{e_{t-\tau}-e_{t}}\right)^{1/\tau}\\
  &=1-\lim_{t\to +\infty}\left[\frac{e_t}{e_{t-\tau}} \times (\frac{e_t-e_{t+\tau}}{e_t})\div (\frac{e_{t-\tau}-e_{t}}{e_{t-\tau}}) \right]^{1/\tau}.
\end{align*}
Since
\begin{align*}
  &\lim_{t\to +\infty}\left(\frac{e_t}{e_{t-\tau}}\right)^{1/\tau}=\lim_{k\to +\infty}\left(\prod_{l=1}^{m}\left(1-\frac{\Delta e_{k+(l-1)\kappa}}{e_{k+(l-1)\kappa}}\right)\right)^{1/(m\kappa)}=C^{1/\kappa},\\
  &\lim_{t\to +\infty}\left(\frac{e_t-e_{t+\tau}}{e_t}\right)^{1/\tau}=\lim_{t\to +\infty}\left(1-\frac{e_{t+\tau}}{e_t}\right)^{1/\tau}=(1-C^m)^{1/(m\kappa)},\\
  &\lim_{t\to +\infty}\left(\frac{e_{t-\tau}-e_{t}}{e_{t-\tau}}\right)^{1/\tau}=\lim_{k\to +\infty}\left(1-\frac{e_{t-\tau}}{e_t}\right)^{1/\tau}=(1-C^m)^{1/(m\kappa)},
\end{align*}
we know that
$$\lim_{t \to +\infty} ACR^{\dagger}_t =1-C^{1/\kappa}.$$
It has been proved in Theorem \ref{theLimit} that
$$\lim_{t\to +\infty}ACR_{t} =1- \lim_{t\to +\infty} \left(\prod^t_{k=1} \frac{e_k}{e_{k-1}}\right)^{1/t}= 1-C^{1/\kappa},$$
and thus, we get that $\lim_{t \to +\infty} ACR_t =\lim_{t \to +\infty} ACR^{\dagger}_t =1-C^{1/\kappa}$.
 \end{proof}
 
The above theorem proves that the limit of  $ACR_t$ is identical to the limit of $ACR^{\dagger}_t$. If the $f^*$ value is unknown, $ACR^{\dagger}_t$ can be used as a replacement of $ACR_t$.

\section{Theoretical Analysis of Evolutionary Programming}
\label{secEP}
\subsection{Landscape-invariant and  Landscape-adaptive Mutation}
\label{secMutation}
In Section~\ref{subsecEmpirical}, it is observed that the ACR of invariant EP tends to zero, while the ACR of adaptive EP tends to a positive constant. This section analyzes  general EP  (Algorithm~\ref{alg4}) using   landscape-invariant or  landscape-adaptive mutation. We only consider genetic operators which are unchanged over time or have  periodic change. So, the population sequence or a subsequence is a homogeneous Markov chain.  

\begin{algorithm}[ht]
\caption{Evolutionary Programming}
\label{alg4}
\begin{algorithmic}[1]
\STATE counter $t \leftarrow 0$;
\STATE initialize $\mu$ individuals $X_0 = (\mathbf{x}_1,\cdots, \mathbf{x}_{\mu})$;
\WHILE{the stopping criterion is not satisfied}
\STATE generate $Y_t=(\mathbf{y}_1,\cdots, \mathbf{y}_{\mu})$ by mutation $Y_t = X_t+Z_t$ where $Z_t =(\mathbf{z}_1,\cdots, \mathbf{z}_{\mu})$ is $\mu$ random variables; if $\mathbf{y}_i$ is out of the definition domain $\mathcal{D}$, let $\mathbf{y}_i=\mathbf{x}_i$;
\STATE evaluate each individual's fitness in population $X_t \cup Y_t$;
\STATE  $X_{t+1}\leftarrow$ select $\mu$ individuals from $X_t \cup Y_t$, where the best individual is always selected;
\STATE counter $t\leftarrow t+1$;
\ENDWHILE
\end{algorithmic}
\end{algorithm}

Mutation $Y=X+Z$  can be characterized by probability transition. Given a population ${X} \in \mathcal{S}$ and a subset $\mathcal A \subset \mathcal{S}$, the  {transition probability kernel} $P_g({X}; \mathcal A)$~\cite{Meyn1993Markov} is defined as
 $$
 P_g({X}; \mathcal A) =\int_{\mathcal A} p_g({X}; Y) dY ,
 $$ where  $p_g({X}; Y)$ is  the probability density function depicting the mutation transition from $ {X}$ to $ {Y}$.  Furthermore, the  $\kappa$-step kernel is defined as 
$$
 P^{(\kappa)}_g({X}; \mathcal A) =\int_{Y \in \mathcal A} p^{(\kappa)}_g( X_{t+\kappa} = Y \mid X_t=X ) dY.
 $$ 
 In this paper, we  assume it is  continuous and bounded.  
 
Generally,   mutation operators can be classified into two categories.
\begin{itemize}
\item \textbf{Landscape-invariant mutation}:  mutation $Y=X+Z$ is called \emph{landscape-invariant} if
 $Z$ is a  random variable vector whose joint probability density function $p_g({0}; {Z})$ is independent on $X$. For simplicity, denote $p_g({0}; {Z})$ as $p_z(Z)$. EP using invariant mutation is named \emph{invariant EP}. For example,  the Gaussian mutation using invariant $\sigma$ in Section~\ref{subsecEmpirical}  belongs to invariant mutation.

\item \textbf{Landscape-adaptive mutation}: if  the probability density function of $Z$ varies on $X$, mutation $Y=X+Z$ is called \emph{landscape-adaptive}. EP using adaptive mutation is named \emph{adaptive EP}. For example,  the Gaussian mutation using adaptive $\sigma$ in Section~\ref{subsecEmpirical} is  adaptive mutation.
\end{itemize}

Given a contraction factor $\rho \in (0,1]$ and a population $X$, the population set $\mathcal{S}$ can be divided into two   disjoint  subsets:
\begin{align}
\label{px}
    \mathcal{S}(X,\rho)=\{ {Y}\in \mathcal{S}|e({Y})< \rho  e(X)\},\quad
  \overline{ \mathcal{S}}(X,\rho)=   \mathcal{S} \setminus \mathcal{S}(X,\rho).
\end{align}
The set $\mathcal{S}(X,\rho)$ is named as a  \emph{$\rho$-promising region}. When $\rho=1$, the set $\mathcal{S}(X,1)$ is called a  \emph{promising region}.

\subsection{Analysis of Landscape-invariant EP}
For EP using landscape-invariant mutation, we prove that its ACR converges to 0.
First we demonstrate that the infinum of the transition probability to the promising region is zero under a mild condition.
\begin{lemma}
\label{lemma2}
If  the number of optimal solutions is finite,  then the transition probability of the landscape-invariant EP to the promising region satisfies
\begin{equation}
    \label{equInvInf}
    \inf \{ P_g(X, S(X,1)); X \notin \mathcal{S}^*\}=0.
\end{equation}
\end{lemma}

\begin{proof}
In order to prove (\ref{equInvInf}),  it is sufficient to prove $\lim_{e(X)\to 0}P_g(X, S(X,1))=0$.
 That is,
$\forall \varepsilon>0, \exists \delta>0$,  $\forall X \in \mathcal{A}(\mathcal{S}^*,\delta) \setminus \mathcal{S}^*$ (where the set $\mathcal{A}(\mathcal{S}^*,\delta)=\{X; e(X) \le \delta \}$), it holds
\begin{equation}
\label{equUpper}
P_g(X, \mathcal{S}(X,1))<\varepsilon.
\end{equation}

For a  Lebesgue-measurable set $\mathcal{A} \subset  \mathcal{S}$, let $m(\mathcal{A})$ denote its Lebesgue measure.
Because $p_z(Z)$ is continuous and bounded, the probability of $X+Z$ falling in a small area is small for a fixed $X$. That is, $\forall \varepsilon>0$, $\exists \delta'>0$ (set $\delta'=\varepsilon/\sup p_z(Z)$),     it holds $\forall \mathcal{A} \subset \mathcal{S}: m(\mathcal{A}) \le \delta'$ and $\forall X \in \mathcal{S}$,
\begin{equation}
    \label{equXplusZ}
    \Pr(X+Z \in \mathcal{A}) =\int_{Z: X+Z \in \mathcal{A}} p_z(X+Z) dZ < \varepsilon.
\end{equation}

Because  the number of optimal solutions is finite (then $m(\mathcal{S}^*)=0$) and $f$ is continuous, for the set $\mathcal{A}(\mathcal{S}^*,\delta)$,  we can choose $\delta$  sufficiently small so that $m(\mathcal{A}(\mathcal{S}^*, \delta)) \le \delta'$.

Because $f$ is continuous,  we may set $\delta$  sufficiently small so
that   $\forall X \in \mathcal{A}(\mathcal{S}^*, \delta)$ and $Y \notin \mathcal{A}(\mathcal{S}^*, \delta)$:  $f(X)<f(Y)$. This implies that $\mathcal{S}(X,1) \subset  \mathcal{A}(\mathcal{S}^*, \delta)$. According to (\ref{equXplusZ}) and $m(\mathcal{A}(\mathcal{S}^*, \delta)) \le \delta'$, $\forall X \in \mathcal{A}(\mathcal{S}^*, \delta)\setminus \mathcal{S}^*$,   we have
\begin{equation*}
    \label{equZ}
    \Pr(X+Z \in \mathcal{A}(\mathcal{S}^*, \delta) )<\varepsilon.
\end{equation*}
Because $\mathcal{S}(X,1) \subset  \mathcal{A}(\mathcal{S}^*, \delta)$, we have
\begin{align*}
    P_g(X, S(X,1))  \le \Pr (X+Z \in \mathcal{A}(\mathcal{S}^*, \delta))  < \varepsilon.
\end{align*}
Then we get (\ref{equUpper}), and the proof is completed.
\end{proof}

The theorem below analyzes the limit property of  invariant EP.
\begin{theorem}
\label{theInvariant}
If the number of optimal solutions is finite   and the invariant EP using elitist selection  converges  in mean, then starting from any $X_0$, $\lim_{t \to +\infty} ACR_t=0$.
\end{theorem}

\begin{proof}
According to Theorem~\ref{theLimit}, it is sufficient to prove
$\lim_{t\to+\infty} \Delta e_t/e_{t-1} =0.$
That is
$\forall \varepsilon>0, \exists t_0>0$, $\forall t \ge t_0$,
\begin{equation}
\label{equInv}
\Delta e_t<\varepsilon e_{t}.
\end{equation}

From  (\ref{equUpper}) in Lemma~\ref{lemma2}, we know $\forall \varepsilon>0,$ $\exists \delta>0$, let $\mathcal{A}(\mathcal{S}^*,\delta)=\{X; e(X) \le \delta \}$. Then $\forall X \in \mathcal{A}(\mathcal{S}^*,\delta) \setminus \mathcal{S}^*$, it holds
\begin{equation}
\label{equInfZero}
P_g(X, \mathcal{S}(X,1))<\varepsilon.
\end{equation}
Since the sequence $\{e_t; t=0, 1, \cdots \}$ converges to 0, EP converges almost surely to 0, that is, $$\Pr(\lim_{t \to +\infty} e(X_t)=0)=1.$$ Denote
$$\mathcal{S}_{1}=\{\omega\in\mathcal{S}|\lim_{t \to +\infty} e(X_t(\omega))=0\},\,\,\mathcal{S}_{2}=\{\omega\in\mathcal{S}|\lim_{t \to +\infty} e(X_t(\omega))\neq 0\}.$$
It is obvious that
\begin{equation}\label{zeroPro}
\Pr(\omega\in\mathcal{S}_2)=0,
\end{equation}
and for the given $\delta >0$, $\exists\,t_0>0$, then $\forall\,t>t_0$, it holds
$$e(X_{t}(\omega))<\delta,\quad \forall\,\omega\in\mathcal{S}_1.$$
From (\ref{equInfZero}) we know
\begin{equation*}
    P_g(X, \mathcal{S}(X_{t}(\omega),1))\le \varepsilon,\quad \forall \omega\in\mathcal{S}_1.
\end{equation*}
Then   we obtain
\begin{equation}
\label{equInfZero1}
\mathbb{E}[e(X_{t}(\omega))-e(X_{t+1}(\omega))\mid X_{t}(\omega)] \le   \varepsilon  e(X_{t}(\omega)),\quad \forall \omega\in\mathcal{S}_1 .
\end{equation}
While $\forall \omega\in\mathcal{S}_2$, we know  there exists a positive $B$:
\begin{equation}
\label{equInfZero2}
\mathbb{E}[e(X_{t}(\omega))-e(X_{t+1}(\omega))\mid X_{t}(\omega)] \le   B.
\end{equation}

Combining (\ref{zeroPro}),  (\ref{equInfZero1}) and (\ref{equInfZero2}) together, we get
\begin{align*}
 \Delta e_t
=&\int_{\mathcal{S}_1}\mathbb{E}[e(X_{t}(\omega))-e(X_{t+1}(\omega))\mid X_{t}(\omega)]\Pr(d\omega)   +\int_{\mathcal{S}_2}\mathbb{E}[e(X_{t}(\omega))-e(X_{t+1}(\omega))\mid X_{t}(\omega)]\Pr(d\omega)\nonumber\\
\le &\varepsilon\int_{\mathcal{S}_1}e(X_{t}(\omega))\Pr(d\omega)+B \cdot 0
\le \varepsilon e_{t}.
\end{align*}
So (\ref{equInv}) is true, and we   complete the proof.
\end{proof}

Theorem \ref{theInvariant} states that for the invariant elitist EP,   $ACR_t \to 0$ as $t \to +\infty$. This means that  landscape-invariant mutation is less efficient in  continuous optimization.  This phenomenon can be explained by the lazy convergence for general Markov search~\cite{tarlowski2017convergence}. As the population approaches the optimum, the probability of generating a better state from one step to another goes to zero. This causes the slow convergence.

Note:   Theorem~\ref{theInvariant}  does not hold if the number of the optimal set ${X}^*$ is not finite, for example, the Jump function~(\ref{equJUMP}) in Section~\ref{secNecssity}.

\subsection{Analysis of  Landscape-adaptive EP}
Landscape-adaptive mutation can be further split into two categories according to the probability of locating promising regions.
\begin{itemize}
\item  \textbf{Zero landscape-adaptive mutation}:   adaptive mutation $Y=X+Z$ is called \emph{zero landscape-adaptive} if
 the transition probability to the promising region satisfies
\begin{equation}
\label{equZero}
\inf \{ P_g(X; \mathcal{S}(X,1)); X \notin \mathcal{S}^*\} =0.
\end{equation}
  
    \item  \textbf{Positive landscape-adaptive mutation}:    adaptive mutation $Y=X+Z$ is called \emph{positive landscape-adaptive} if $\exists \rho \in (0,1)$, the transition probability to the $\rho$-promising region satisfies
\begin{equation}
\label{thecon}
C_{\rho}= \inf \{ P^{(\kappa)}_g(X; \mathcal{S}(X,\rho)); X \notin \mathcal{S}^*\} >0.
\end{equation}
for some positive integer $\kappa$.
\end{itemize}
Thus, landscape-adaptive EP can be classified into two categories: the \emph{zero landscape-adaptive EP} employing  zero landscape-adaptive mutation and  the \emph{positive landscape-adaptive EP} with  positive landscape-adaptive mutation.
Theorems \ref{theZeroPositive} and \ref{thePositiveAdaptive} analyze the limit property of the ACR of the two types of EP.

\begin{theorem}\label{theZeroPositive}
If  the number of optimal solutions is finite and the zero landscape-adaptive EP using elitist selection converges in mean, then starting from some $X_0$, the ACR satisfies $\lim_{t \to +\infty} ACR_t =0.$
\end{theorem}

\begin{proof}
For zero landscape-adaptive mutation, (\ref{equZero}) implies that there exists a subsequence $\{X_{t'}\}$ such that   $\lim_{t'\to +\infty}P_g(X_{t'}, S(X_{t'},1))=0$.  Similar to the proof of Theorem~\ref{theInvariant}, we know that  $\lim_{t \to +\infty} ACR_{t'}=0$.
\end{proof}

Theorem \ref{theZeroPositive} states that for the zero landscape-adaptive elitist EP, its ACR  tends to 0  when starting from some initial population. Thus, zero landscape-adaptive mutation is not always efficient. It is different from Theorem \ref{theInvariant} which holds for any initial non-optimal population.    

\begin{theorem}\label{thePositiveAdaptive}
If  the number of optimal solutions is finite and the positive landscape-adaptive EP using elitist selection converges in mean, then starting from any non-optimal population $X_0$,
the ACR  satisfies
$\varliminf_{t \to +\infty} ACR_t \ge C>0$.
\end{theorem}

\begin{proof}
Let us estimate the lower bound of the ACR limit. 
From (\ref{px}), we know that for any $k \ge 0$,
\begin{align*}
\mathcal{S}(X_{k},\rho)&=\{Y \in  \mathcal{S}\mid e(Y) \le \rho e(X_{k})\}.
\end{align*}
It follows that $\mathcal{S}(X_{k},\rho)\subset \mathcal{S}(X_{k},1)$, and for any $ Y \in\mathcal{S}(X_{k},\rho)$,
\begin{align}
  f(X_{k})-f( Y )\ge (1-\rho) (f(X_{k})-f^*).\label{equMinusF}
\end{align}
So we get for $\kappa$ consecutive steps that
\begin{align}
  & e(X_k)-e(X_{k+\kappa})\nonumber\\
   = & \int_{\mathcal{S}(X_{k},1)} (f(X_{k})-f( Y)) p^{(\kappa)}_g( X_{t+\kappa} = Y \mid X_t=X )d Y\nonumber\\
   \ge & \int_{\mathcal{S}(X_{k},\rho)} (f(X_{k})-f( Y)) p^{(\kappa)}_g( X_{t+\kappa} = Y \mid X_t=X )d Y\nonumber\\
  \ge &  \int_{\mathcal{S}(X_{k},\rho)} (1-\rho) \left(f(X_{k})-f^*\right) p^{(\kappa)}_g( X_{t+\kappa} = Y \mid X_t=X )d Y\quad \mbox{(from (\ref{equMinusF}))}\nonumber\\
  \ge & (1-\rho) C_{\rho} e(X_{k}) . \quad \mbox{(from (\ref{thecon}))}\label{equCondExp}
\end{align}
Then
\begin{align}
    \frac{\Delta^{(\kappa)} e_k}{e_{k}}
    \ge  \frac{(1-\rho) C_{\rho}\mathbb{E}[ e(X_{k})]}{e_{k}}  =   (1-\rho) C_{\rho}. \label{LB}
\end{align}
It holds that
\begin{align*}
\varliminf_{t \to +\infty}ACR_t= \varliminf_{t \to +\infty} \left[1-\left( \frac{e_t}{e_{0}} \right)^{1/t}\right]= \varliminf_{t \to +\infty} \left[ 1- \left(\prod_{k=1}^{t} \frac{e_k}{e_{k-1}} \right)^{1/t}\right] \ge (1-\rho) C_{\rho}=C.
\end{align*}
Let $C=(1-\rho) C_{\rho}>0$. We complete the proof.
\end{proof}

Theorem \ref{thePositiveAdaptive} proves that positive landscape-adaptive mutation is efficient because it results in a positive lower bound of the ACR. Theorems~\ref{theInvariant} to~\ref{thePositiveAdaptive} explain the different behaviors between the landscape-adaptive EP and landscape-invariant EP. To design an efficient EA, mutation should be positive landscape-adaptive.

\subsection{Case Study of Landscape-invariant and Landscape-adaptive EP}
 This section makes a case study  of EP using landscape-adaptive and landscape-invariant mutation. 
Consider the following minimization problem:
\begin{align}
\label{equAbstract}
\min f_A(\mathbf x)=\|\mathbf x\|_{\infty}=\max\{|x_1|,|x_2|\},\quad \mathbf x=(x_1,x_2)\in\mathbb R^2.
\end{align}
It is optimized by  $(1+1)$ EP using mutation $\mathbf{y}= \mathbf{x}+ \boldsymbol{z\Gamma}$, where $\mathbf{z}$ is a random vector subject to a uniform distribution $\mathcal U([-1,1]\times[-1,1])$. $\boldsymbol\Gamma=diag\{\gamma_1,\gamma_2\}$ represents the step sizes along $x_1$ and $x_2$ axes respectively. Denote the individual at the $t^{th}$ generation as $\mathbf x_t=(x_1,x_2)$. Due to symmetry of the landscape, we can postulate without loss of generality that $x_1\ge x_2$. Then, $\|\mathbf x_t\|_{\infty}=x_1$.

\textbf{Scenario 1:   $\boldsymbol\Gamma$ is constant.} In this case, the uniform mutation is landscape-invariant. Theorem~\ref{theInvariant} implies that   $ACR_t$ converge to $0$ when $t \to +\infty$.

We first validate that this landscape-invariant (1+1) EP converges in probability. Let $\mathbf{x}_t=(x_1,x_2)$, set $$\gamma_1=|x_1 |,\quad \gamma_2=|x_2 |.$$
$\forall\,0<\delta<\min\{\gamma_1,\gamma_2\}$, denote $$\mathcal R_{\delta}=[-\delta,\delta]\times [-\delta,\delta].$$
The elitist selection confirms that
\begin{align*}  
  \Pr(\mathbf{x}_{t+1}\notin\mathcal R_{\delta}|\mathbf{x}_t\notin\mathcal R_{\delta})\le\Pr(\mathbf{x}_1\notin\mathcal R_{\delta}|\mathbf{x}_0\notin\mathcal R_{\delta})=\frac{\delta^2}{4\gamma_1\gamma_2} \quad \forall t\in\mathbb Z^+.
\end{align*}
Thus,
\begin{align*}
  \Pr(\mathbf{x}_{t+1}\notin\mathcal R_{\delta})&=\Pr(\mathbf{x}_{t+1}\notin\mathcal R_{\delta}|\mathbf{x}_t\notin\mathcal R_{\delta})\Pr(\mathbf{x}_t \notin\mathcal R_{\delta})\\
  \le & \frac{\delta^2}{4\gamma_1\gamma_2}\Pr(\mathbf{x}_t\notin\mathcal R_{\delta})\le\cdots\le \left(\frac{\delta^2}{4\gamma_1\gamma_2}\right)^{t+1}\Pr(\mathbf{x}^{(0)}\notin\mathcal R_{\delta})
\end{align*}
and we know
\begin{align}\label{ConInPro}
  \lim_{t\to\infty}\Pr(\mathbf{x}_{t+1}\notin\mathcal R_{\delta})=0.
\end{align}
That is, this landscape-invariant (1+1) EP converges in probability.

From (\ref{ConInPro}) we know $\forall \varepsilon_1>0$, there exists $t_0>0$ such that
  \begin{equation}\label{SmallProb}
   \Pr(\mathbf{x}_{t}\notin\mathcal R_{\delta})<\varepsilon_1,\quad t>t_0.
\end{equation}
Meanwhile,  
\begin{align}\label{CondImprovement}
  \mathbb E[e(\mathbf{x}_t)-e(\mathbf{x}_{t+1})|{\mathbf x}_t=(x_1,x_2)]  &=\int_{\mathcal{S}(\mathbf{x}_t, 1)}\frac{1}{4\gamma_1\gamma_2}(x_1-\max\{|y_1|,|y_2|\})dy_1dy_2 \nonumber\\
  &=\frac{2}{4\gamma_1 \gamma_2}\int_{-x_1}^{x_1}\left[\int_{-x_1}^{y_1}(x_1-|y_1|)dy_2\right]dy_1  =\frac{[x_1]^3}{2\gamma_1\gamma_2}.
\end{align}
 
By setting $\epsilon=\frac{\delta^2}{\gamma_1\gamma_2}$, (\ref{SmallProb}) and (\ref{CondImprovement}) imply that
\begin{align*}
  \Delta e_{t}&=\mathbb E[\mathbb E[e(\mathbf{x}_t)-e(\mathbf{x}_{t+1})|{\mathbf x}_{t}=(x_1,x_2)]]\\
  =&\int_{\mathbf{x}_t \in\mathcal R_{\delta}}\frac{[x_1]^3}{2\gamma_1\gamma_2}d\mathbf{P}_{\mathbf{x}_t}+\int_{\mathbf{x}_t \notin\mathcal R_{\delta}}\frac{[x_1]^3}{2\gamma_1\gamma_2}d\mathbf{P}_{\mathbf{x}_t}\\
  <&\frac{\epsilon}{2}\mathbb E[e(\mathbf{x}_t)]+M\epsilon_1\mathbb E[e(\mathbf{x}_t)],
\end{align*}
where $M=\max\{\frac{[x_1]^2}{2\gamma_1\gamma_2}-\frac{\epsilon}{2}\}$. Hence, $\lim_{t\to\infty}\frac{\Delta e_{t}}{e_{t}}\le\frac{\epsilon}{2}$. 
Since $\epsilon$ is arbitrarily small, we can conclude that
\begin{align*}
  \lim_{t\to\infty}ACR_t=& 1-\lim_{t\to\infty}\left(\frac{e_t}{e_0}\right)^{1/t}=0.
\end{align*}

\textbf{Scenario 2:   $\boldsymbol\Gamma$ is  adaptive to the landscape.}
We consider an adaptive mutation strategy which dynamically adjusts the step size as below.  Let $\mathbf{x}_t=(x_1,x_2)$,
\begin{align}\label{S2}
  \gamma_i(t)=\left\{\begin{aligned}&0.1 |x_i|, && 2m\kappa\le t<(2m+1)\kappa,\\
  &0.2 |x_i|, && 2(m+1)\kappa\le t<(2m+2)\kappa, \end{aligned}\quad m\in {\mathbb Z}^+,i=1,2,\right.
\end{align}
where $\kappa$ is a positive integer. 

Because $\frac{\gamma_i(t)}{|x_i|}$ is bounded from below by $0.1$, we have
\begin{align}
        \inf \{ P_g(X; \mathcal{S}(X,1)); X \notin \mathcal{S}^*\}>0.
\end{align}
Thus, the above strategy is  {positive landscape-adaptive}. According to Thereom~\ref{thePositiveAdaptive}, there exists some positive $C$ such that $\lim_{t \to +\infty} ACR_t \ge C$. Now let us deduce the lower and upper bounds on the limit of $ACR_t$. Set
\begin{align*}
  \lambda=\left\{\begin{aligned}&0.1, && 2m\kappa\le t<(2m+1)\kappa; \\
  &0.2, && 2(m+1)\kappa\le t<(2m+2)\kappa, \end{aligned}\quad m\in\mathbb Z,i=1,2.\right.
\end{align*}
The analysis is split into two different cases.
\begin{enumerate}
  \item While $0<(1+\lambda)x_2 <x_1 $, we discuss two cases:  $(1+\lambda)x_2\le (1-\lambda)x_1$ and  $(1-\lambda)x_1<(1+\lambda)x_2$.
\begin{enumerate}
    \item If $(1+\lambda)x_2\le (1-\lambda)x_1$, we know  
  \begin{align*}
       \mathbb E[e(\mathbf{x}_{t})-e(\mathbf{x}_{t+1})|{\mathbf x}_{t}=(x_1,x_2)]= &\int_{\mathcal{S}(\mathbf{x}_t, 1)}\frac{1}{4\gamma_1\gamma_2}(x_1-\max\{|y_1|,|y_2|\})dy_1dy_2\\
  = &\frac{1}{4\lambda^2x_1 x_2 }\int_{(1-\lambda)x_1 }^{x_1 }\left[\int_{(1-\lambda)x_2 }^{(1+\lambda)x_2 }(x_1 -y_1)dy_2\right]dy_1=\frac{\lambda}{4}x_1.
      \end{align*}
      Then,
      \begin{equation}\label{Result1}
        \frac{\Delta e_t}{e_t}=\frac{1}{4}\lambda.
      \end{equation}

\item If $(1-\lambda)x_1<(1+\lambda)x_2$, we know
  \begin{align*}
         \mathbb E[e(\mathbf{x}_t)-e(\mathbf{x}_{t+1})|{\mathbf x}_{t}=(x_1,x_2)]=&\int_{\mathcal{S}(\mathbf{x}_t, 1)}\frac{1}{4\gamma_1\gamma_2}(x_1-\max\{|y_1|,|y_2|\})dy_1dy_2\\
  \le &\frac{1}{4\lambda^2 x_1 x_2} \int_{(1-\lambda) x_1}^{x_1 } \left[\int_{(1-\lambda)x_2}^{(1+\lambda)x_2}(x_1^{(t)}-y_1)dy_2\right]dy_1=\frac{\lambda}{4}x_1,
      \end{align*}
  and
  \begin{align*}
         \mathbb E[e(\mathbf{x}_t - e(\mathbf{x}_{t+1})|{\mathbf x}_t]=(x_1 , x_2]=&\int_{\mathcal{S}(\mathbf{x}_t, 1)}\frac{1}{4\gamma_1\gamma_2}(x_1-\max\{|y_1|,|y_2|\})dy_1dy_2\\
  \ge &\frac{1}{4\lambda^2x_1 x_2 }\int_{(1-\lambda)x_1 }^{x_1 }\left[\int_{(1-\lambda)x_2 }^{(1-\lambda)x_1 }(x_1 -y_1)dy_2\right]dy_1=\frac{1}{8}\lambda(1-\lambda)x_1 .
      \end{align*}
      Then,
      \begin{equation}\label{Result2}
        a_1(\lambda) \le \frac{\Delta e_t}{e_t} \le b_1(\lambda),
      \end{equation}
      where
      \begin{equation*}
        a_1(\lambda)=\frac{1}{8}\lambda(1-\lambda),\qquad b_1(\lambda)=\frac{1}{4}\lambda.
      \end{equation*}
\end{enumerate}

  \item While $(1+\lambda)x_2 \ge x_1 \ge x_2 $, we know
  \begin{align*}
         \mathbb E[e(\mathbf{x}_t)-e(\mathbf{x}_{t+1})|{\mathbf x}_t]=(x_1,x_2)]=&\int_{\mathcal{S}(\mathbf{x}_t, 1)} \frac{1}{4\gamma_1\gamma_2}(x_1 -\max\{|y_1|,|y_2|\})dy_1dy_2\\
  =&\frac{1}{4\lambda^2x_1 x_2 }\int_{(1-\lambda)x_1 }^{x_1 }\left[\int_{(1-\lambda)x_2 }^{x_1 }(x_1 -\max\{|y_1|,|y_2|\})dy_2\right]dy_1\\  =&\frac{1}{4\lambda^2x_1 x_2 }\left[(\frac{1}{2}-\frac{1}{6}\lambda)(x _1 )^3-\frac{1}{2}(1-\lambda)(x_1 )^2 x_2 \right]\\ =&\frac{x_1 }{4}\left[(\frac{1}{2}-\frac{1}{6}\lambda) \frac{x_1 }{x_2 }-\frac{1}{2}(1-\lambda)\right].
      \end{align*}
      Then,
      \begin{align*}
        \frac{1}{12}\lambda x_1  \le \mathbb E[e(\mathbf{x}_t)-e(\mathbf{x}_{t+1})|{\mathbf x}_t]=(x_1 ,x_2 )] \le \frac{1}{6}\lambda(5-\lambda)x_1,
      \end{align*}
      and it holds
      \begin{equation}\label{Result3}
        a_2(\lambda) \le \frac{\Delta e_t}{e_t} \le b_2(\lambda),
      \end{equation}
      where
      \begin{equation*}
        a_2(\lambda)=\frac{1}{12}\lambda,\qquad b_2(\lambda)=\frac{1}{6}\lambda(5-\lambda).
      \end{equation*}
\end{enumerate}

In summary, (\ref{Result1}), (\ref{Result2}) and (\ref{Result3}) imply  
\begin{equation*}
  \min\{\frac{1}{8}\lambda(1-\lambda),\frac{1}{12}\lambda\}\le\frac{\Delta e_t}{e_t}\le \frac{1}{6}\lambda(5-\lambda).
\end{equation*}
Since
\begin{equation*}
  ACR_t=1-\left(\prod_{k=0}^{t-1}(1-\frac{\Delta e_k}{e_k})\right)^{1/t},
\end{equation*}
the ACR is bounded by
\begin{equation*}
  \min\{\frac{1}{8}\lambda(1-\lambda),\frac{1}{12}\lambda\}\le ACR_t\le \frac{1}{6}\lambda(5-\lambda).
\end{equation*}
Set $\lambda=0.1$ or $\lambda=0.2$. Thus, the ACR of the landscape-adaptive EP bounded by
\begin{equation}
  \frac{1}{120}\le ACR_t\le \frac{4}{25}.
\end{equation}

\section{Relationship between ACR and Decision Space Dimension}
\label{secRelation} 
\subsection{Polynomial ACR versus Exponential ACR}
In Section~\ref{subsecNumRelation}, computer simulation demonstrates that the ACR decrease as the the decision space dimension increases. In theory, Teytaud and Selly~\cite{teytaud2006general} discussed the convergence by the logarithm of the distance to the optimum and proved a linear convergence with constant in $[1-O(1/d),1]$ for the comparison-based algorithms. In general, the relationship between the ACR and decision space dimension can be studied similar to the time complexity.
Recall in the theory of time complexity, the runtime of  algorithms is often classified into two categories: exponential runtime and polynomial runtime as the problem input size. Similarly,  the  ACR of  EAs can be classified into polynomial and exponential categories as follows.

\begin{itemize}
  \item \textbf{Polynomial ACR}: starting from any $X_0$,  for any $t$,  $ACR_t$ is not less than a \emph{reciprocal of a polynomial function  of $d$}, that is
  \begin{equation}
  \label{equPolynomial}
  \exists a>1,  \forall t\ge 0: ACR_t = \Omega\left(d^{-a}\right),
  \end{equation}
where $\Omega$ is under Bachmann-Landau notation. (\ref{equPolynomial}) means $ACR_t$ reduces slowly  as $d$ increases.  

  \item \textbf{Exponential ACR}: starting from some $X_0$, for a  period  as long as an exponential function of $d$, $ACR_t$ is not more than a \emph{reciprocal of an exponential  function of $d$}, that is
  \begin{equation}
  \label{equExponential}
   \exists a>1, b>1,  \forall t = \Omega(b^d): ACR_t = O \left(a^{-d}\right).
  \end{equation}
 where $O$ is under Bachmann-Landau notation. (\ref{equExponential}) means $ACR_t$ reduces quickly as $d$ increases.
\end{itemize}

According to the definition, a polynomial ACR is a linear ACR. However, an exponential ACR could be either  a sublinear or linear ACR. For example, an exponential ACR such that $\lim_{t \to +\infty} ACR_t (d) = \exp(-d)$ is still a linear ACR.

The theorem  below provides  sufficient conditions  of determining whether  an ACR is polynomial and exponential.

\begin{theorem}\label{theRelationship}
(1) If starting from any $X_0$,  for any $t\ge 0$, some $a>1$ and $\kappa>0$, the rate of $\kappa$-step {error change}  satisfies
\begin{align}
\label{equPoly}
  \frac{\Delta^{(\kappa)} e_t}{e_t} =  \Omega(d^{-a}).
\end{align}
then starting from any $X_0$, for any $t\ge 0$,   $ ACR_t = \Omega\left(d^{-a/\kappa}\right).$

(2) If   starting from some $X_0$, for some $t = O(b^d)$ where $b>1$, some $a>1$ and and $\kappa>0$, the rate of {error change} satisfies
\begin{align}
\label{equExpo}
  \frac{\Delta^{(\kappa)} e_t}{e_t} = O \left(a^{-d}\right).
\end{align}
then starting from the above $X_0$,   for some $t = O(b^d)$,
$ACR_t  = O \left(a^{-d/\kappa}\right).
$
\end{theorem}

\begin{proof}
Denote $t=m\kappa+k$, where $m$ is a positive integer, $k\in\{0,1,\dots,\kappa-1\}$. Because
\begin{align*}
  ACR_t
  =1-\left[\prod_{k=1}^{t}\frac{ e_{k}}{e_{k-1}}\right]^{1/t}=1-\left[\prod_{l=0}^{m-1}\left(1-\frac{\Delta^{(\kappa)} e_{l\kappa}}{e_{l\kappa}}\right)\frac{ e_{m\kappa+k}}{e_{m\kappa}}\right]^{1/(m\kappa+k)},
\end{align*}
from (\ref{equPoly}), we derive $  ACR_t = \Omega(d^{-a/\kappa})$.  
From (\ref{equExpo}), we have for some $t = O(b^d)$, $  ACR_t = O(d^{-a/\kappa})$.
\end{proof}
In the above theorem, two parameters $a$ and $b$ relies on problems and algorithms.

\subsection{Polynomial ACR on Easy Problems}\label{secLinear}
To exemplify a polynomial ACR on easy problems, consider the  (1+1) adaptive RUS  for minimizing the linear function.
\begin{equation}\label{equLinear}
  \min f_L(\mathbf{x})=\textstyle \sum_{i=1}^{d} c_{i} x_i, \quad \mathbf{x}=(x_1,\dots,x_d)\in [0,1]^d \subset \mathbb R^d.
\end{equation}
$f_L$ is a natural extension of linear functions from pseudo-Boolean function optimization~\cite{he2001drift} to continuous optimization. The OneMax function (\ref{equOneMax})  is a special instance of linear functions. Since the OneMax function is easiest to the adaptive (1+1) EP, the ACR slowly decreases as $d$ on the OneMax function. Let us make a rigorous analysis of this claim on linear functions.

Let  $\mathbf{x}_t=\mathbf{x}=(x_1,\dots,x_d)$ be a non-optimal solution. An offspring $\mathbf{y}$ is generated by search along a randomly selected dimension $j$ with probability ${1}/{d}$. That is,
$$\mathbf{y}_j=(x_1,\dots,x_{j-1},y_{j},x_{j+1},\dots,x_d),$$
where $y_{j}=x_{j}+\mathcal N(0,\sigma_j)$. Thanks to elitist selection, $\mathbf{y}$ is accepted if and only if it satisfies $y_j<x_j$.  
It is trivial to validate that the probability of hitting the promising region is
\begin{align}
   P_g(\mathbf{x};\mathcal{S}(\mathbf{x},1)) =\sum_{j=1}^{d}\frac{1}{\sqrt{2\pi}\sigma_jd} \int_{0}^{x_j}e^{-\frac{( y-x_j)^2}{2\sigma_j^2}}d y=\frac{1}{d}\sum_{j=1}^{d}\left(\frac{1}{2}-\Phi\left(-\frac{x_j}{\sigma_j}\right)\right). \label{equPromising2}
  \end{align}

Let $\sigma_j$ such that $x_j/\sigma_j=C_0$ where $C_0>0$.
In this case, we prove that the Gaussian mutation is positive landscape-adaptive, that is, $\exists C>0$, $\rho\in (0,1)$, $\forall \mathbf{x} \notin  {X}^*$,
$
P_g(\mathbf{x};\mathcal{S}(\mathbf{x},\rho))\ge C .
$

Since $x_j/\sigma_j=C_0 >0$, we have
\begin{align}
  P_g(\mathbf{x};\mathcal{S}(\mathbf{x},1))=\frac{1}{2}-\Phi(-C_0).
  \end{align}
 Take $P_g(\mathbf{x},\mathcal{S}(\mathbf{x},\rho))$ as a function of $\rho$ defined in the interval $(0,1]$. Obviously  $P_g(\mathbf{x},\mathcal{S}(\mathbf{x},\rho))$ is continuous, that is, $\forall\,\,\varepsilon>0$,   $\exists \rho(\varepsilon) \in (0,1)$ such that
 \begin{equation}\label{Cont}
 P_g(\mathbf{x},\mathcal{S}(\mathbf{x},\rho(\varepsilon)))>P_g(\mathbf{x},\mathcal{S}(\mathbf{x},1))-\varepsilon.
 \end{equation}
Let $\rho=\rho(\varepsilon)$ and $C_{\rho}=\frac{1}{2}-\Phi(-C_0)-\varepsilon$ for any given $\frac{1}{2}-\Phi(-C_0)>\varepsilon>0$. (\ref{Cont}) and (\ref{thecon}) confirm that the Gaussian mutation with $x_j/\sigma_j=C_0$ is positive landscape-adaptive. Then, Theorem \ref{thePositiveAdaptive} claims that the ACR is not less than a positive,
 \begin{equation}\label{LbAcr}
 ACR_t \ge C=(1-\rho(\varepsilon))(\frac{1}{2}-\Phi(-C_0)-\varepsilon).
 \end{equation}

However, the lower bound presented by (\ref{LbAcr})  does not show how the ACR is connected to the  dimension $d$. In the following, we demonstrate a relationship between the ACR and decision space dimension. 

Given a fixed $j$, for the   Gaussian mutation with $x_j/\sigma_j=C_0$, the  error change  is
\begin{align}\label{555}
  &  \frac{1}{\sqrt{2\pi}\sigma_j}\int_{0}^{x_j} c_j(x_j-y)\exp\left\{-\frac{(y-x_j)^2}{2\sigma_j^2}\right\}dy =\frac{\sigma_j c_j}{\sqrt{2\pi}}\left(1-e^{-\frac{x_j^2}{2\sigma_j^2}}\right)=\frac{\sigma_j c_j}{\sqrt{2\pi}}\left(1-e^{-\frac{C_0^2}{2}}\right).
\end{align} 
Since $j$ is chosen at random from $\{1, \cdots, d\}$, the average error change over all   $j=1, \cdots, d$ is
\begin{align*}
 \Delta e_t= \frac{1}{d}\sum_{j=1}^d \frac{\sigma_j c_j}{\sqrt{2\pi}}\left(1-e^{-\frac{C_0^2}{2}}\right)=\frac{1}{\sqrt{2\pi}dC_0}\left(1-e^{-\frac{C_0^2}{2}}\right)\sum_{j=1}^d c_jx_j.
\end{align*} 
Then we know that the rate of error change is
\begin{align*} 
 \frac{\Delta e_{t}}{e_t}=&\frac{1}{d}\frac{1}{\sqrt{2\pi}C_0}\left(1-e^{-\frac{C_0^2}{2}}\right) =\Theta(\frac{1}{d}).
\end{align*} 
According to Theorem \ref{theRelationship},  $ACR_t$ is the  reciprocal of a polynomial function  of $d$ for any $t$.

\subsection{Exponential ACR on Hard Problems}
To exemplify an exponential ACR on hard problems, consider the (1+1) adaptive EP for minimizing the  deceptive function (\ref{equdeceptive}). Since the deceptive function is hard to the adaptive (1+1) EP, the ACR quickly decreases as $d$ on the deceptive function. Let us make a rigorous analysis of this claim.

Let $\mathbf{x}_t =\mathbf{x}$ be a non-optimal solution. By Gaussian mutation, an offspring $\mathbf{y}$ is generated by $\mathbf{x}$ as
$y_i={x}_i+\mathcal N({0}, {\sigma}_i)$ where $i=1, \cdots, d$.
Let $ \sigma_i =x_i$ where $i=1, \cdots, d$.
We prove that this Gaussian mutation is positive landscape-adaptive.

For the deceptive function, let (1+1) EP start  from the local optimum $(1, \cdots, 1)$.
According to the definition of the deceptive function (\ref{equdeceptive}),  $f_D(\mathbf{y}) < f_D(\mathbf{x})$ if and only if $\sum_{i=1}^d y_i \le 1/2$. Thus, only an offspring $\mathbf{y}$ with $\sum_{i=1}^d y_i \le 1/2$ can be accepted, and  the transition probability to the promising region is
\begin{align} 
 P_g(\textbf{x}, \mathcal{S}(\textbf{x},1))=&\int_{\sum_i y_i\le 1/2} p_{\mathbf{x}}(\mathbf{y})d\mathbf{y}\nonumber\\
=&\int_{\sum_i y_i \le 1/2} \frac{1}{(\sqrt{2\pi})^d}\exp\left\{-\frac{\sum_i(y_i-1)^2}{2}\right\}dy_1\dots dy_d\nonumber\\
\le & \int_{\mathbf{y}\in [0,1/2]^d} \frac{1}{(\sqrt{2\pi})^d}\exp\left\{-\frac{\sum_i(y_i-1)^2}{2}\right\}dy_1\dots dy_d  \le b^{-d},
\end{align}
for some $b>1$.

The expected hitting time to the promising region is at least $b^d$. We choose a constant $c \in (0,1)$ such that $1< c b <b$, then for $t \le (cb)^d$, the event of $\mathbf{x}_t$ leaving the local optimum happens with a probability at most $1-(1-b^{-d})^{(cb)^d}\le c^{d}$. Thus  $e_t =d- O(c^{d})$.

For any $\mathbf{x}_t$  satisfying $\sum_{i=1}^d x_i \le 1/2$, its average error change   is at most $c^{d}/2 $. For $\mathbf{x}_t =(1,\cdots,1)$, its average error change is
\begin{align} 
&\int_{\sum_i y_i\le 1/2}\left((d+1-\sum_i x_i)-\sum_i y_i\right)p_{\mathbf{x}}(\mathbf{y})d\mathbf{y}\nonumber\\
=&\int_{\sum_i y_i \le 1/2}\left(1-\sum_i y_i \right)\frac{1}{(\sqrt{2\pi})^d}\exp\left\{-\frac{\sum_i(y_i-1)^2}{2}\right\}dy_1\dots dy_d\nonumber\\
\le & \int_{\mathbf{y}\in [0,1/2]^d}\left(1-\sum_i y_i \right)\frac{1}{(\sqrt{2\pi})^d}\exp\left\{-\frac{\sum_i(y_i-1)^2}{2}\right\}dy_1\dots dy_d\nonumber\\
=&\int_{\mathbf{z}\in [-1/2,0]^d}\left(1-\sum_i (1+z_i)\right)\frac{1}{(\sqrt{2\pi})^d}\exp\left\{-\frac{\sum_iz_i^2}{2}\right\}dz_1\dots dz_d\nonumber\\
\le & \int_{\mathbf{z}\in [-1/2,0]^d}\frac{1}{(\sqrt{2\pi})^d}\exp\left\{-\frac{\sum_iz_i^2}{2}\right\}dz_1\dots dz_d\nonumber\\
= &  \left(\Phi(0)-\Phi(-1/2)\right)^d. \label{999}
\end{align}
Then, by setting $a=1/(\Phi(0)-\Phi(-1/2)) $, we know that $a>1$. Then the rate of error change is not more than the  reciprocal of an {exponential} function of $d$, that is
\begin{align}
\frac{\Delta e_t}{e_t}\le \frac{a^{-d}+O(c^{d})}{d-O(c^{d})}.
\end{align}

According to Theorem \ref{theRelationship},  we conclude that  $ACR_t$ is not more than the  reciprocal of an {exponential} function of $d$ for $t \le (cb)^d$.  This theoretical result confirms $ACR_t$    decreases quickly as $d$ on the deceptive function.

For the (1+1) adaptive RUS on the deceptive function, it is trivial to prove that starting  from the local optimum $\mathbf{x}_0=(1,\cdots, 1)$, the algorithm cannot generate a better child $\mathbf{y}$ such that $f_D(y) < f_D(1,\cdots, 1)$. Thus, for any $t$, we have $\mathbf{x}_t =\mathbf{x}_0$  and then   $ACR_t=0$.

\section{Conclusions}
\label{secConclusions}
This paper conducts a theoretical analysis of the ACR of EAs in continuous optimization. According to the limit property, the ACR is classified into two categories: (1) linear ACR whose limit inferior value is larger than a positive  and (2) sublinear ACR whose value  converges to zero. Then, it is proven that for EP using  positive landscape-adaptive mutation, its ACR is linear. But for EP using landscape-invariant or zero landscape-adaptive mutation, its ACR is sublinear.    

The relation  between the ACR  and  the decision space dimension is also classified into two categories: (1) polynomial ACR whose value is larger than the  reciprocal of a polynomial function of the dimension for any generation, and (2) exponential ACR whose value is less than the  reciprocal of an exponential function of the dimension for an exponential long period. It is proven that for easy functions such as  linear functions, the ACR of the (1+1) adaptive RUS  is polynomial. But for hard functions such as the deceptive function,  the ACR of both the (1+1) adaptive EP and RUS is exponential. 

This paper does not discuss EAs whose genetic operators change over time. This topic will be left for future research.


\end{document}